\documentclass{article} 
\usepackage{iclr2026_conference,times}

\usepackage{amsmath,amsfonts,bm}

\def\eqref#1{equation~\ref{#1}}

\def\1{\bm{1}}

\DeclareMathAlphabet{\mathsfit}{\encodingdefault}{\sfdefault}{m}{sl}
\SetMathAlphabet{\mathsfit}{bold}{\encodingdefault}{\sfdefault}{bx}{n}

\usepackage{comment}
\usepackage[colorlinks=true, urlcolor=blue, linkcolor=red, citecolor=darkgreen]{hyperref}
\usepackage{url}
\usepackage{graphicx}
\usepackage{multicol}
\usepackage{multirow}
\usepackage{colortbl}

\def\0{{\bf 0}}
\def\1{{\bf 1}}

\definecolor{purple}{rgb}{0.56,0.27,0.68}
\definecolor{red}{rgb}{0.95,0.4,0.4}
\definecolor{purered}{rgb}{1,0,0}
\definecolor{blue}{rgb}{0.4,0.4,0.95}
\definecolor{darkblue}{rgb}{0,0,0.8}
\definecolor{grey}{rgb}{0.6,0.6,0.6}
\definecolor{col1}{RGB}{232, 161, 148}
\definecolor{col11}{RGB}{255, 228, 228}
\definecolor{col2}{RGB}{148, 187, 232}
\definecolor{col33}{RGB}{206, 239, 255}
\definecolor{col3}{RGB}{233, 255, 245}
\definecolor{lightgrey}{rgb}{0.85,0.85,0.85}
\definecolor{lightlightgrey}{rgb}{0.9,0.9,0.9}
\definecolor{verylightBG}{rgb}{0.9,0.99,0.99}
\definecolor{darkgreen}{rgb}{0., 0.85, 0.5}
\definecolor{lightgray}{gray}{0.75}

\definecolor{gtred}{RGB}{204, 0, 0}
\definecolor{predgreen}{RGB}{31, 237, 31}
\definecolor{figGreen}{RGB}{56, 118, 29}
\definecolor{lightgray}{gray}{0.75}

\usepackage{algorithm2e}
\usepackage{pythonhighlight} 
\usepackage{fontawesome5}
\lstnewenvironment{pythonic}[1][]{\lstset{style=mypython, frame=none, #1}}{}

\usepackage[capitalize]{cleveref}
\crefname{section}{Sec.}{Secs.}
\Crefname{section}{Section}{Sections}
\Crefname{table}{Table}{Tables}
\crefname{table}{Tab.}{Tabs.}

\newcommand{\method}{RF-DETR}

\title{\method: Neural Architecture Search for Real-Time Detection Transformers}

\iclrfinalcopy
\author{Isaac Robinson$^{1}$, Peter Robicheaux$^{1}$, Matvei Popov$^1$, Deva Ramanan$^{2}$, Neehar Peri$^{2}$ \\
$^1$Roboflow, $^2$Carnegie Mellon University\\
}

\begin{document}

\maketitle

\begin{abstract}
Open-vocabulary detectors achieve impressive performance on COCO, but often fail to generalize to real-world datasets with out-of-distribution classes not typically found in their pre-training. Rather than simply fine-tuning a heavy-weight vision-language model (VLM) for new domains, we introduce \method, a light-weight specialist detection transformer that discovers accuracy-latency Pareto curves for any target dataset with weight-sharing neural architecture search (NAS). Our approach fine-tunes a pre-trained base network on a target dataset and evaluates thousands of network configurations with different accuracy-latency tradeoffs \textit{without re-training}. Further, we revisit the ``tunable knobs'' for NAS to improve the transferability of DETRs to diverse target domains. Notably, \method \ significantly improves over prior state-of-the-art real-time methods on COCO and Roboflow100-VL. \method \ (nano) achieves 48.0 AP on COCO, beating D-FINE (nano) by 5.3 AP at similar latency, and \method \ (2x-large) outperforms GroundingDINO (tiny) by 1.2 AP on Roboflow100-VL while running $20 \times$ as fast. To the best of our knowledge, \method \ (2x-large) is the first real-time detector to surpass 60 AP on COCO. Our code is available on \href{https://github.com/roboflow/rf-detr}{GitHub}.
\end{abstract}

\section{Introduction}
Object detection is a fundamental problem in computer vision that has matured in recent years~\citep{felzenszwalb2009object, lin2014coco, ren2015faster}. Open-vocabulary detectors like GroundingDINO \citep{liu2023grounding} and YOLO-World \citep{Cheng2024YOLOWorld} achieve remarkable zero-shot performance on common categories like {\tt car}, {\tt truck}, and {\tt pedestrian}. However, state-of-the-art vision-language models (VLMs) still struggle to generalize to out-of-distribution classes, tasks and imaging modalities not typically found in their pre-training \citep{robicheaux2025roboflow100vl}. Fine-tuning VLMs on a target dataset significantly improves in-domain performance at the cost of runtime efficiency (due to heavy-weight text encoders) and open-vocabulary generalization. In contrast, specialist (i.e., closed-vocabulary) object detectors like D-FINE \citep{peng2024dfine} and RT-DETR \citep{zhao2024rtdetr} achieve real-time inference, but underperform fined-tuned VLMs like GroundingDINO. In this paper, we modernize specialist detectors by combining internet-scale pre-training with real-time architectures to achieve state-of-the-art performance {\em and} fast inference. 

\textbf{Are Specialist Detectors Over-Optimized for COCO?} Sustained progress in object detection can be largely attributed to standardized benchmarks like PASCAL VOC \citep{pascalvoc} and COCO \citep{lin2014coco}. However, we find that recent specialist detectors implicitly overfit to COCO at the cost of real-world performance using bespoke model architectures, learning rate schedulers, and augmentation schedulers. Notably, state-of-the-art object detectors like YOLOv8 \citep{yolov8} generalize poorly to real-world datasets with significantly different data distributions from COCO (e.g., number of objects per image, number of classes, and dataset size). To address these limitations, we present \method, a scheduler-free approach that leverages internet-scale pre-training to generalize to real-world data distributions. To better specialize our model for diverse hardware platforms and dataset characteristics, we revisit neural architecture search (NAS) in the context of end-to-end object detection and segmentation.

\textbf{Rethinking Neural Architecture Search (NAS) for DETRs.} NAS discovers accuracy-latency tradeoffs by exploring architectural variants within a pre-defined search space. NAS has been previously studied in the context of image classification \citep{tan2019efficientnet, cai2019ofa} and for model sub-components like detector backbones \cite{tan2020efficientdet} and FPNs \cite{ghiasi2019fpn}. Unlike prior work, we explore \textit{end-to-end} weight-sharing NAS for object detection and segmentation. Our key insight, inspired by OFA \citep{cai2019ofa}, is that we can vary model inputs like image resolution, and architectural components like patch size during training. Further, weight-sharing NAS allows us to modify inference configurations like the number of decoder layers and query tokens to specialize our strong base model \textit{without fine-tuning}. We evaluate all model configurations with grid search on a validation set. Importantly, our approach does not evaluate the search space until the base model has been fully-trained on the target dataset. As a result, all possible sub-nets (i.e., model configurations within the search space) achieve strong performance without further fine-tuning, significantly reducing the computational cost of optimizing for new hardware. Interestingly, we find that sub-nets not explicitly seen during training still achieve high performance (Appendix \ref{appendix:tunable-knobs}), suggesting that \method \ can generalize to unseen architectures. Extending \method \ for segmentation is also relatively straightforward and only requires adding a lightweight instance segmentation head. We denote this model as \method-Seg. Notably, this allows us to also leverage end-to-end weight-sharing NAS to discover Pareto optimal architectures for real-time instance segmentation.

\textbf{Standardizing Latency Evaluation.} We evaluate our approach on COCO \citep{lin2014coco} and Roboflow100-VL (RF100-VL) \citep{robicheaux2025roboflow100vl} and achieve state-of-the-art performance among real-time detectors. \method \ (nano) outperforms D-FINE (nano) by 5\% AP on COCO at comparable run-times, and \method \ (2x-large) beats GroundingDINO (tiny) on RF100-VL at a fraction of the runtime. \method-Seg (nano) outperforms YOLOv11-Seg (x-large) on COCO while running 4 $\times$ as fast. However, comparing \method's latency with prior work remains challenging because reported latency evaluation varies significantly between papers. Notably, each new model re-benchmarks the latency of prior work for fair comparison on their hardware. For example, D-FINE's reported latency evaluation of LW-DETR \citep{chen2024lw} is 25\% faster than originally reported. We identify that this lack of reproducibility can be primarily attributed to GPU power throttling during inference. We find that buffering between forward passes limits power over-draw and standardizes latency evaluation (Table \ref{tab:latency}).

\begin{figure}[t]
    \centering
    \includegraphics[trim={0cm 0cm 0cm 0cm},clip,width=0.47\linewidth]{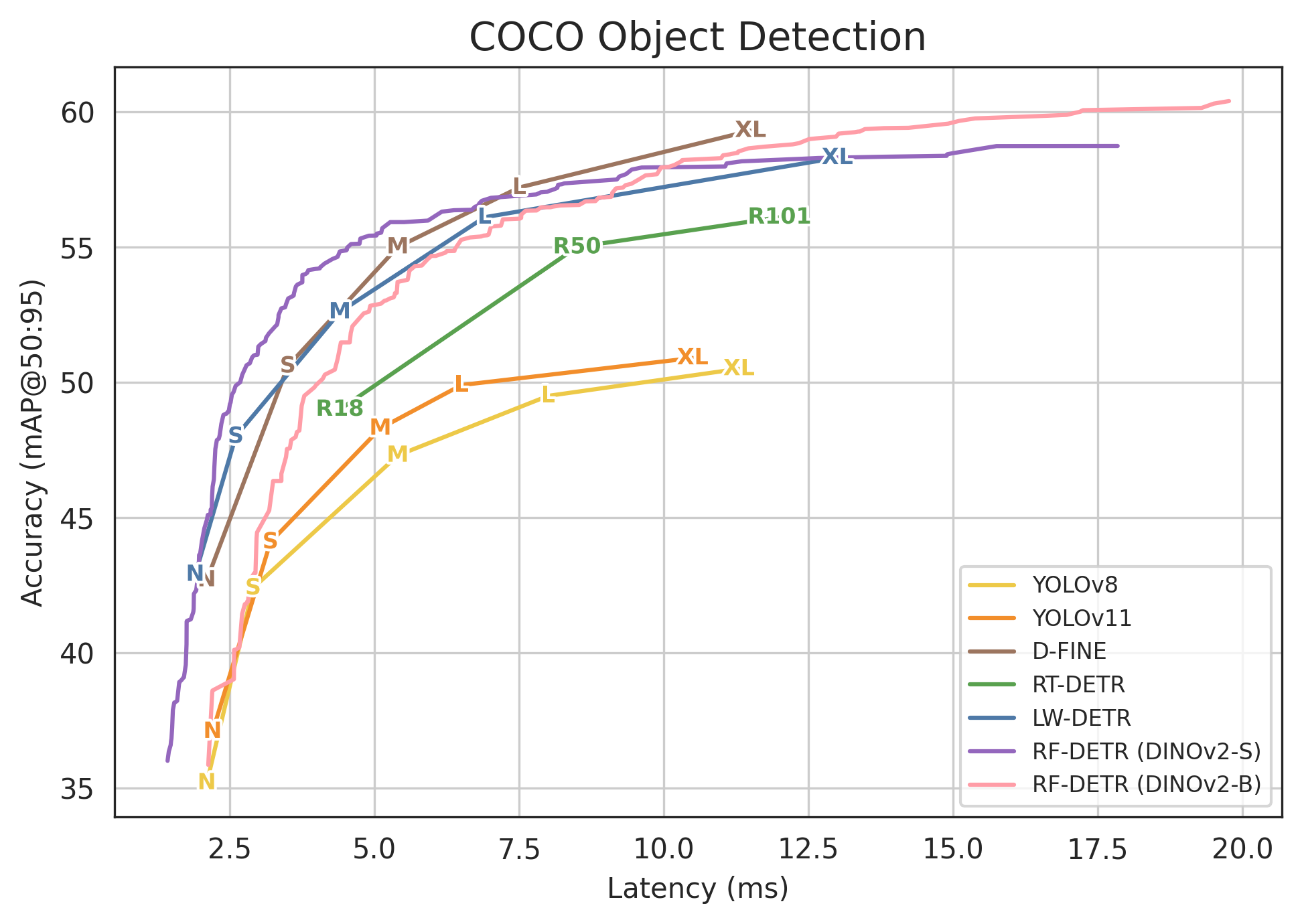}
    \includegraphics[trim={0cm 0cm 0cm 0cm},clip,width=0.47\linewidth]{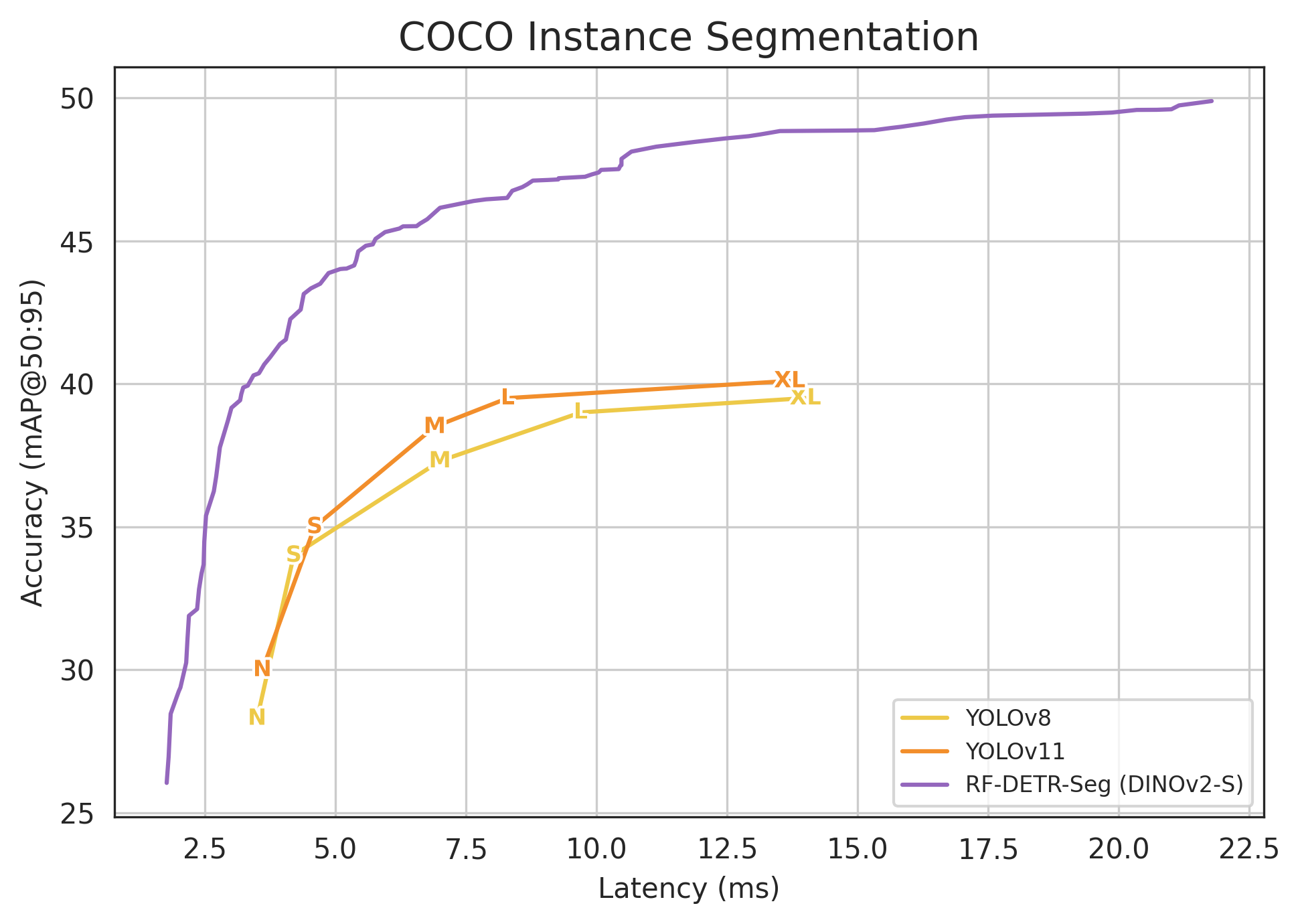} \\
    \includegraphics[trim={0cm 0cm 0cm 0cm},clip,width=0.47\linewidth]{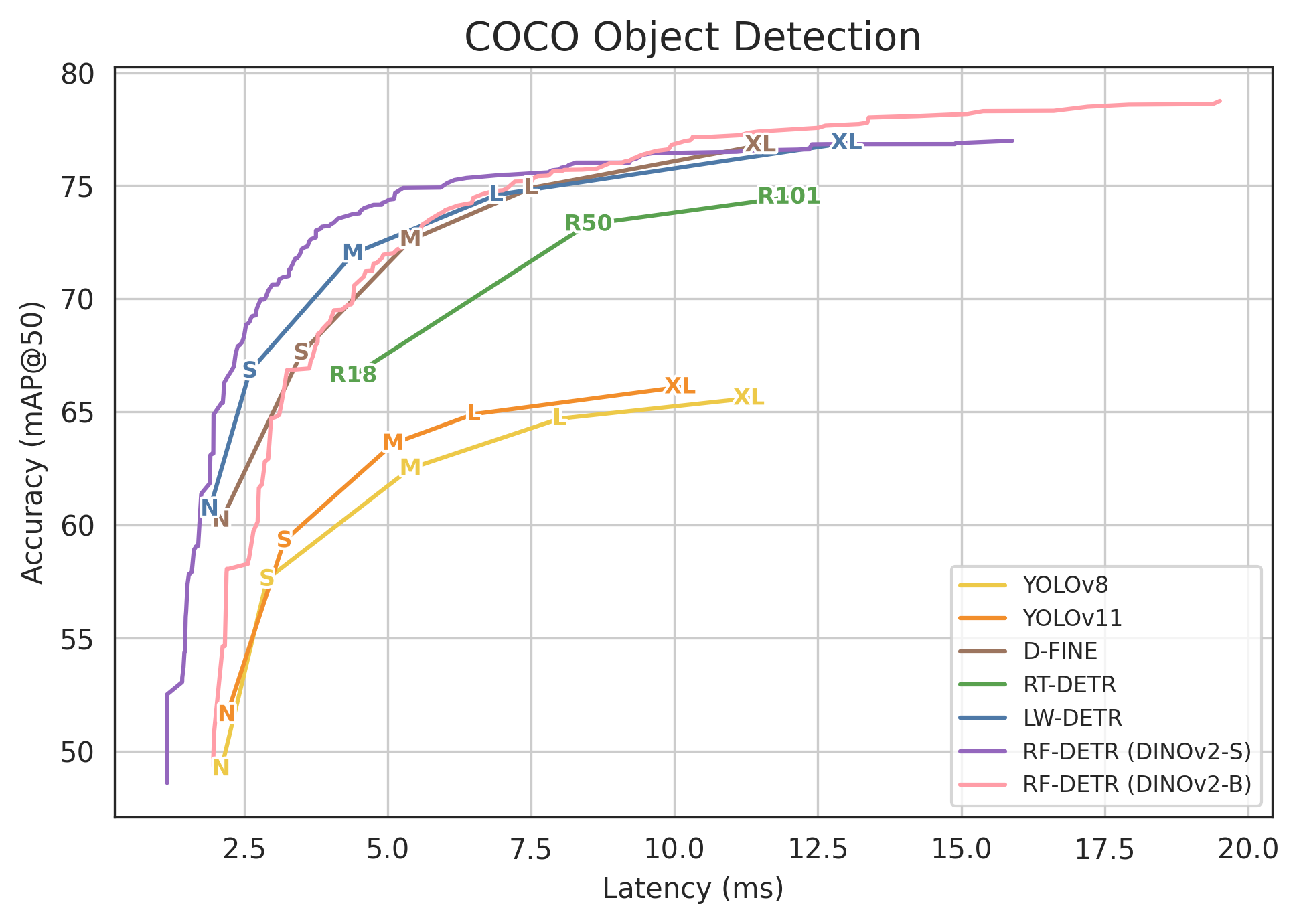}
    \includegraphics[trim={0cm 0cm 0cm 0cm},clip,width=0.47\linewidth]{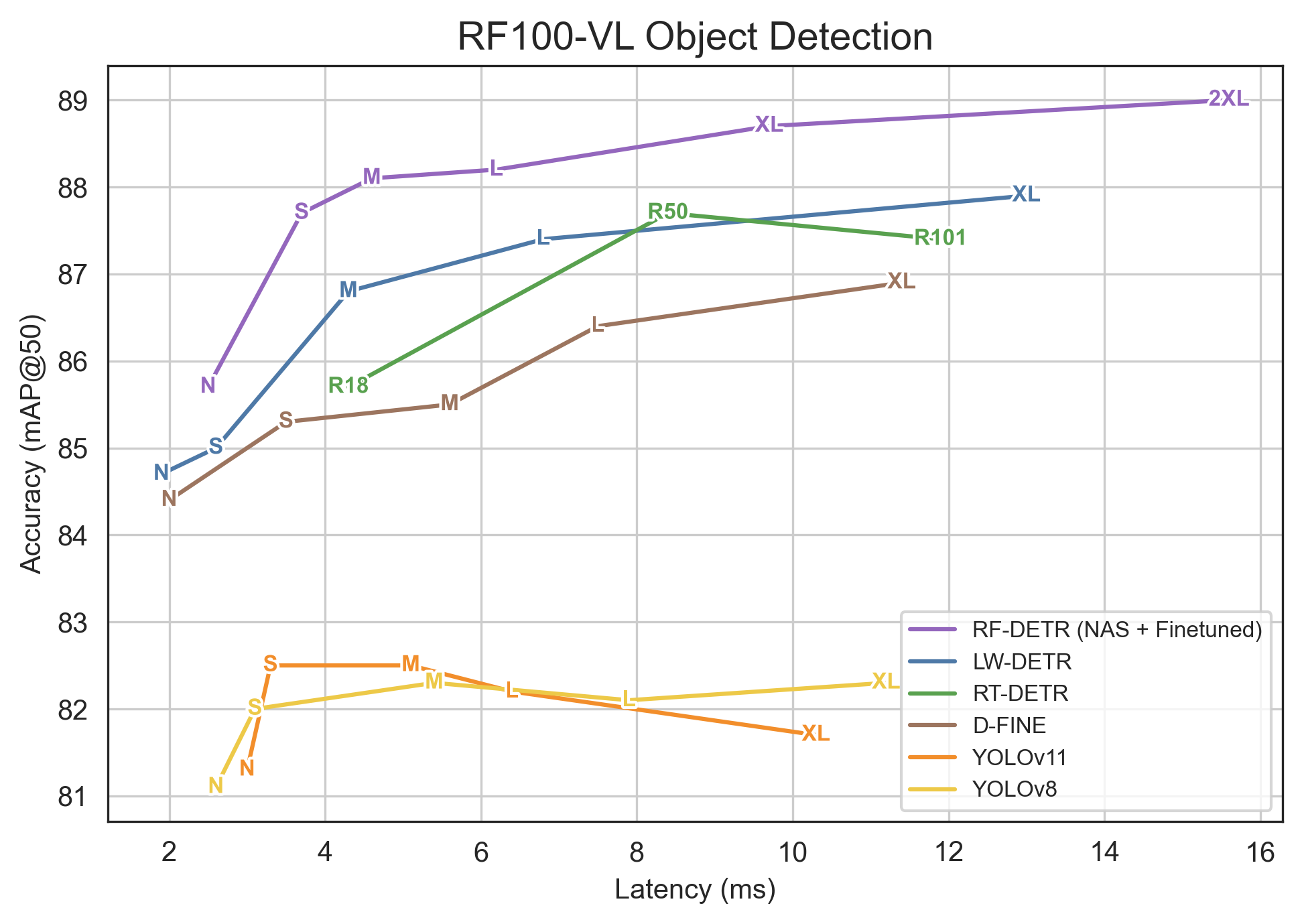}

    \caption{{\bf Accuracy-Latency Pareto Curve.} We plot the Pareto accuracy-latency frontier for real-time detectors on the COCO detection val-set (top left, bottom left), COCO segmentation val-set (top right), and RF100-VL test-set (bottom right). Since RF100-VL contains 100 distinct datasets, we select target latencies for the N, S, M, L, XL, 2XL configurations, search for \method \ models with latencies within 10\% of the target and report their average performance after fine-tuning to convergence. Importantly, all points along \method's continuous Pareto curves for COCO are derived from a single training run.}
    \label{fig:pareto-curve}
\end{figure}

\textbf{Contributions.} We present three major contributions. First, we introduce \method, a family of scheduler-free NAS-based detection and segmentation models that outperform prior state-of-the-art on RF100-VL \citep{robicheaux2025roboflow100vl} and real-time methods with latencies $\leq$ 40 ms on COCO \citep{lin2014coco} (Figure \ref{fig:pareto-curve}). To the best of our knowledge, \method \ is the first real-time detector to exceed 60 mAP on COCO. Next, we explore the ``tunable-knobs'' for weight-sharing NAS to improve accuracy-latency tradeoffs for end-to-end object detection (Figure \ref{fig:nas-knobs}). Notably, our use of a weight-sharing NAS allows us to leverage large-scale pre-training and effectively transfer to small datasets (Table \ref{tab:rf100-vl}). Lastly, we revisit current benchmarking protocols for measuring latency and propose a simple standardized procedure to improve reproducibility.

\section{Related Works}

\textbf{Neural Architecture Search (NAS)} automatically identifies families of model architectures with different accuracy-latency tradeoffs \citep{zoph2016neural, zoph2018learning, real2019regularized, cai2018efficient}. Early NAS approaches \citep{zoph2016neural, real2019regularized} focused primarily on maximizing accuracy, with little consideration for efficiency. As a result, discovered architectures (e.g., NASNet and AmoebaNet) were often computationally expensive. More recent hardware-aware NAS methods \citep{cai2018proxylessnas, tan2019mnasnet, wu2019fbnet} address this limitation by incorporating hardware feedback directly into the search process. However, these methods must repeat the search and training process for each new hardware platform. In contrast, OFA \citep{cai2019ofa} proposes a weight-sharing NAS that decouples training and search by simultaneously optimizing thousands of sub-nets with different accuracy-latency tradeoffs. Contemporary methods typically evaluate NAS for object detection by simply replacing standard backbones with NAS backbones in existing detection frameworks. Unlike prior work, we directly optimize end-to-end object detection accuracy to find Pareto optimal accuracy-latency tradeoffs for any target dataset.

\textbf{Real-Time Object Detectors} are of significant interest for safety-critical and interactive applications. Historically, two-stage detectors like Mask-RCNN \citep{he2017mask} and Hybrid Task Cascade \citep{chen2019hybrid} achieved state-of-the-art performance at the cost of latency, while single-stage detectors like YOLO \citep{redmon2016you} and SSD \citep{liu2016ssd} traded accuracy for state-of-the-art runtime. However, modern detectors~\citep{zhao2024rtdetr} reexamine this accuracy-latency tradeoff, simultaneously improving on both axes. Recent YOLO variants innovate on architecture, data augmentation, and training techniques \citep{redmon2016you,wang2023yolov7,wang2024yolov9,yolov8,yolov11} to improve performance while maintaining fast inference. Despite their efficiency, most YOLO models rely on non-maximum suppression (NMS), which introduces additional latency. In contrast, DETR \citep{carion2020end} removes hand-crafted components like NMS and anchor boxes. However, early DETR variants \citep{zhu2020deformable, zhang2022dino, meng2021conditional, liu2022dab} achieved strong accuracy at the cost of runtime, limiting their use in real-time applications. Recent works such as RT-DETR \citep{zhao2024rtdetr} and LW-DETR \citep{chen2024lw} have successfully adapted high performance DETRs for real-time applications. Building on LW-DETR, \method \ is the first real-time detector to achieve more than 60 AP on COCO. 

\textbf{Vision-Language Models} are trained on large-scale, weakly supervised image-text pairs from the web. Such internet-scale pre-training is a key enabler for open-vocabulary object detection \citep{liu2023grounding, Cheng2024YOLOWorld}. GLIP \citep{li2021grounded} frames detection as phrase grounding with a single text query, while Detic \citep{zhou2022detecting} boosts long-tail detection using ImageNet-level supervision \citep{russakovsky2015imagenet}. MQ-Det \citep{xu2024multi} extends GLIP with a learnable module that enables multi-modal prompting. Recent VLMs demonstrate strong zero-shot performance and are often applied as “black-box’’ models in diverse downstream tasks \citep{ma2023long, pmlr-v205-peri23a, khurana2024shelf, sal2024eccv, takmaz2025cal}. However, \citet{robicheaux2025roboflow100vl} find that such models perform poorly when evaluated on categories not typically found in their pre-training, requiring further fine-tuning. In addition, many vision-language models are prohibitively slow, making them difficult to use for real-time tasks. In contrast, \method \ combines the fast inference of real-time detectors with the internet-scale priors of VLMs to achieve state-of-the-art performance on RF100-VL and at all latencies $\leq$ 40 ms on COCO.

\section{\method: Weight-Sharing NAS With Foundation Models}
\begin{figure}[t]
    \centering
    \includegraphics[width=0.95\linewidth]{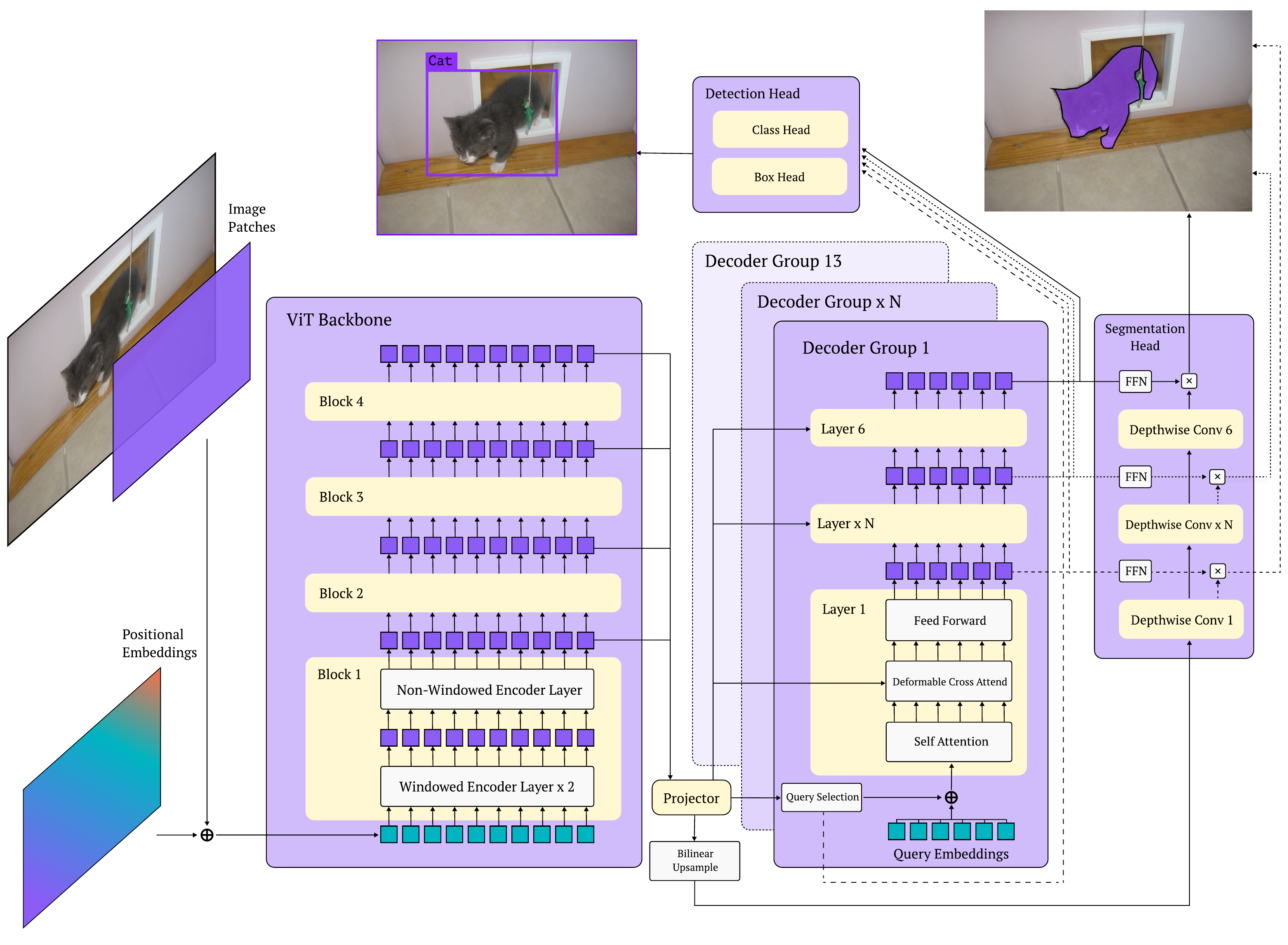}
    \caption{{\bf \method \ Architecture}. \method \ uses a pre-trained ViT backbone to extract multi-scale features of the input image. We interleave windowed and non-windowed attention blocks to balance accuracy and latency. Notably, the deformable cross-attention layer and segmentation head both bilinearly interpolate the output of the projector, allowing for consistent spatial organization of features. Lastly, we apply detection and segmentation losses at all decoder layers to facilitate decoder drop out at inference.}
    \label{fig:arch}
\end{figure}

\begin{figure}[t]
    \centering
    \includegraphics[width=0.95\linewidth]{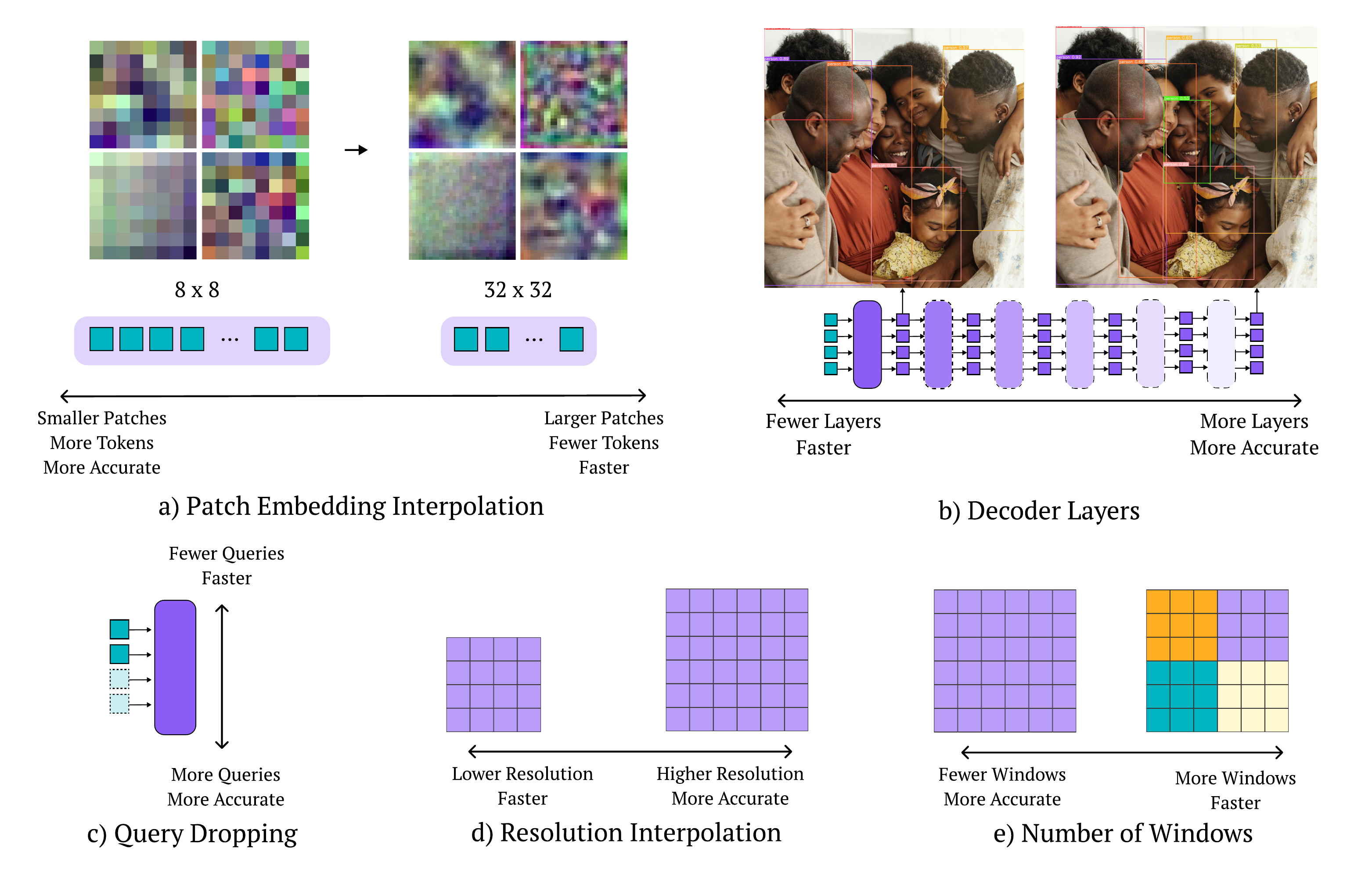}
    \caption{{\bf NAS Search Space}. We vary (a) patch size, (b) number of decoder layers, (c) number of queries, (d) image resolution, and (e) number of windows per attention block when evaluating different operating points along \method's Pareto curve. In addition to training thousands of network configurations in parallel, we find that this ``architecture augmentation'' serves as a regularizer and improves generalization.}
    \label{fig:nas-knobs}
\end{figure}

In this section, we describe the architecture of our base model (Figure \ref{fig:arch}) and present the ``tunable knobs'' of our weight-sharing NAS (Figure \ref{fig:nas-knobs}). Further, we highlight the limitations of hand-designed learning-rate and augmentation schedulers, and advocate for a scheduler-free approach.

\textbf{Incorporating Internet-Scale Priors.} 
\method \ modernizes LW-DETR \citep{chen2024lw} by simplifying its architecture and training procedure to improve generalization to diverse target domains. First, we replace LW-DETR's CAEv2 \citep{zhang2022cae} backbone with DINOv2 \citep{oquab2023dinov2}. We find that initializing our backbone with DINOv2's pre-trained weights significantly improves detection accuracy on small datasets. Notably, CAEv2's encoder has 10 layers with a patch size of 16, while DINOv2's encoder has 12 layers. 
Our DINOv2 backbone has more layers and is slower than CAEv2, but we make up for this latency using NAS (discussed next). Lastly, we facilitate training on consumer-grade GPUs via gradient accumulation by using layer norm instead of batch norm in the multi-scale projector. 

\textbf{Real-Time Instance Segmentation.} Inspired by \cite{li2023maskdino}, we add a lightweight instance segmentation head to jointly predict high quality segmentation masks. Our segmentation head bilinearly interpolates the output of the encoder and learns a lightweight projector to generate a pixel embedding map. Specifically, we upsample the same low-resolution feature map for the detection and segmentation heads to ensure that it contains relevant spatial information. Unlike MaskDINO \citep{li2023maskdino}, we do not incorporate multi-scale backbone features in our segmentation head to minimize latency. Lastly, we compute the dot product of all projected query token embeddings (at the output of each decoder layer transformed by a FFN) with the pixel embedding map to generate segmentation masks.  Interestingly, we can interpret these pixel embeddings as segmentation prototypes \citep{bolya2019yolact}. Motivated by LW-DETR's observation that pre-training improves DETRs, we pre-train \method-Seg on Objects-365 \citep{objects365} psuedo-labeled with SAM2 \citep{ravi2024sam2} instance masks. 

\textbf{End-to-End Neural Architecture Search.}
Our weight-sharing NAS evaluates thousands of model configurations with different input image resolutions, patch sizes, window attention blocks, decoder layers, and query tokens. At every training iteration, we uniformly sample a random model configuration and perform a gradient update (Appendix \ref{sec:implementation_details}). This allows our model to efficiently train thousands of sub-nets in parallel, similar to ensemble learning with dropout \citep{dropout}. We find that this weight-sharing NAS approach also serves as a regularizer during training, effectively performing ``architecture augmentation''. To the best of our knowledge, \method \ is the first end-to-end weight-sharing NAS applied to object detection and segmentation. We describe each component below.

\begin{itemize}
  \item  \textit{Patch Size}. Smaller patches lead to higher accuracy at greater computational cost. We adopt a FlexiVIT-style \citep{beyer2023flexivit} transformation to interpolate between patch sizes during training. 

  \item  \textit{Number of Decoder Layers.} Similar to recent DETRs \citep{peng2024dfine, zhao2024rtdetr}, we apply a regression loss to the output of all decoder layers during training. Therefore, we can drop any (or all) decoder blocks during inference. Interestingly, removing the entire decoder during inference effectively turns \method \ into a single-stage detector. Notably, truncating the decoder also shrinks the size of the segmentation branch, allowing for greater control over segmentation latency.

  \item  \textit{Number of Query Tokens.} Query tokens learn spatial priors for bounding box regression and segmentation. We drop query tokens (ordered by the maximum sigmoid of the corresponding class logit per token at the output of the encoder, see Appendix \ref{appendix:query_decoder}) at test time to vary the maximum number of detections and reduce inference latency. The Pareto optimal number of query tokens implicitly encodes dataset statistics about the average number of objects per image in a target dataset. 
  
  \item  \textit{Image Resolution.} Higher resolution improves small object detection performance, while lower resolution improves runtime. We pre-allocate $N$ positional embeddings corresponding to the largest image resolution divided by the smallest patch size and interpolate these embeddings for smaller resolutions or larger patch sizes. 
    
  \item  \textit{Number of Windows per Windowed Attention Block}. Window attention restricts self-attention to only process a fixed number of neighboring tokens. We can add or remove windows per block to balance accuracy, global information mixing,  and computational efficiency.
\end{itemize}

At inference time, we pick a specific model configuration to select an operating point on the accuracy-latency Pareto curve. Importantly, different model configurations may have similar parameter counts but significantly different latencies. Similar to \cite{cai2019ofa}, we see little benefit from fine-tuning the NAS-mined models on COCO (Appendix \ref{appendix:fine-tuning}), but note modest improvements from fine-tuning NAS-mined models on RF100-VL. This additional fine-tuning is optional, and is often unnecessary for practical deployment. We posit that \method \  benefits from additional fine-tuning on RF100-VL because the ``architecture augmentation'' regularization requires more than 100 epochs to converge on small datasets. Notably, prior weight-sharing NAS methods \citep{cai2019ofa} train in stages and use a different learning-rate scheduler per-stage. However, such schedulers make strict assumptions about model convergence, which may not hold across diverse datasets. 

\textbf{Training Schedulers and Augmentations Bias Model Performance.}
State-of-the-art detectors often require careful hyper-parameter tuning to maximize performance on standard benchmarks. However, such bespoke training procedures implicitly bias the model towards certain dataset characteristics (e.g. number of images). Concurrent with DINOv3 \citep{simeoni2025dinov3}, we observe that cosine schedules assume a known (fixed) optimization horizon, which is impractical for diverse target datasets like those in RF100-VL. 
Data augmentations introduce similar biases by presuming prior knowledge of dataset properties. For example, prior work leverages aggressive data augmentation (e.g., {\tt VerticalFlip}, {\tt RandomFlip}, {\tt RandomResize}, {\tt RandomCrop}, {\tt YOLOXHSVRandomAug}, and {\tt CachedMixUp}) to increase effective dataset size. However, certain augmentations like {\tt VerticalFlip} may negatively bias model predictions in safety-critical domains. For example, a {\tt person} detector in a self-driving vehicle should not be trained with {\tt VerticalFlip} to avoid false positive detections from reflections in puddles. Therefore, we limit augmentations to horizontal flips and random crops. 
Lastly, LW-DETR applies a per-image random resize augmentation, where each image is padded to match the largest image in the batch. As a result, most images have significant padding, which introduces window artifacts, and wastes computation on padded regions. In contrast, we resize images at the batch level to minimize the number of padded pixels per-batch and to ensure that all positional encoding resolutions are equally likely to be seen at train time.

\section{Experiments}
We evaluate \method \ on COCO and RF100-VL and demonstrate that our approach achieves state-of-the-art accuracy among all real-time methods. In addition, we identify inconsistencies in standard benchmarking protocols and present a simple standardized procedure to improve reproducibility. Following LW-DETR \citep{chen2024lw}, we group models of similar latency into the same size bucket rather than grouping based on parameter count.

\textbf{Datasets and Metrics.} We evaluate \method \ on COCO for fair comparison with prior work and on RF100-VL to evaluate generalization to real-world datasets with significantly different data distributions. Due to the diversity of RF100-VL's 100 datasets, we posit that overall performance on this benchmark is a proxy for transferability to any target domain. We use pycocotools to report standard metrics like mean average precision (mAP) and provide breakdown analysis for AP$_{50}$, AP$_{75}$, AP$_{Small}$, AP$_{Medium}$, and AP$_{Large}$. Further, we evaluate efficiency by measuring GFLOPs, number of parameters, and inference latency on an NVIDIA T4 GPU with Tensor-RT 10.4 and CUDA 12.4. 

{
\setlength{\tabcolsep}{1.6em}
\begin{table}[t]
\caption{\textbf{Standardizing Latency Evaluation.} Variance in latency measurements can be largely attributed to power throttling and GPU overheating. We mitigate this issue by buffering for 200ms between forward passes. Notably, this benchmarking approach is not designed to measure sustained throughput, but rather ensures reproducible latency measurements. We are unable to reproduce YOLOv8 and YOLOv11's mAP results in TensorRT, likely because these models evaluate with multi-class NMS but only use single-class NMS in inference. We use the standard NMS-tuned confidence threshold of 0.01. YOLOv8 and YOLOv11 performance degrades further when quantizied from FP32 to FP16, reaffirming that all models should report latency and accuracy using the same model artifact. Notably, naively quantizing D-FINE to FP16 reduces performance to 0.5 AP. We fix this issue by changing the authors' export code to use ONNX opset 17 (Appendix \ref{sec:implementation_details}).
} 
\label{tab:latency}
\scalebox{0.7}{
\begin{tabular}{|l|cc|cc|cc|cc|}
\hline
\rowcolor{gray!10}
\multicolumn{1}{|c|}{\multirow{1}{*}{\textbf{Method}}} & \multicolumn{2}{c|}{\textbf{Reported}} & \multicolumn{2}{c|}{\textbf{Buffering (FP-32)}} & \multicolumn{2}{c|}{\textbf{Buffering (FP-16)}}  \\
\rowcolor{gray!10}
\multicolumn{1}{|c|}{}                                 & \multicolumn{1}{c|}{AP$_{50:95}$}  & Latency (ms)  & \multicolumn{1}{c|}{AP$_{50:95}$}       & Latency (ms)     & \multicolumn{1}{c|}{AP$_{50:95}$}       & Latency (ms)     \\ \hline
YOLOv8 (M)                                               & \multicolumn{1}{c|}{50.2}       &    5.86      & \multicolumn{1}{c|}{49.3}            &      14.8        & \multicolumn{1}{c|}{47.3}            &   5.4  \\ \hline
YOLOv11 (M)                                                 & \multicolumn{1}{c|}{51.5}       &    4.7      & \multicolumn{1}{c|}{49.7}            &        18.7      & \multicolumn{1}{c|}{48.3}         &    5.2  \\ \hline \hline
RT-DETR (R18)                                                 & \multicolumn{1}{c|}{49.0}       &       4.61   & \multicolumn{1}{c|}{49.0}            &        12.2      & \multicolumn{1}{c|}{49.0}         &  4.4   \\ \hline
LW-DETR (M)                                                 & \multicolumn{1}{c|}{52.5}       &       5.6   & \multicolumn{1}{c|}{52.6}            &        26.8      & \multicolumn{1}{c|}{52.6}         &  4.4   \\ \hline
D-FINE (M)                                                  & \multicolumn{1}{c|}{55.1}       &       5.62   & \multicolumn{1}{c|}{55.1}            &       13.9       & \multicolumn{1}{c|}{55.0 (0.5$^*$)}         &   5.4   \\ \hline
\method \ (M)                                                 & \multicolumn{1}{c|}{-}       &       -   & \multicolumn{1}{c|}{54.8}            &       20.5      & \multicolumn{1}{c|}{54.7}         &   4.4   \\ \hline
\end{tabular}
}
\end{table}
}

\textbf{Standardizing Latency Benchmarking.} Despite its maturity, benchmarking object detectors remains inconsistent across prior work. For example, YOLO-based models often omit non-maximal suppression (NMS) when computing latency, leading to unfair comparisons with end-to-end detectors. Additionally, YOLO-based segmentation models measure the latency of generating prototype predictions instead of directly usable per-object masks \citep{yolov11}, leading to biased runtime measurements. Further, D-FINE's reported latency evaluation of LW-DETR is 25\% faster than reported by \cite{chen2024lwdetr}. We observe that such differences can be attributed to detectable power throttling events, particularly when the GPU overheats (Table \ref{tab:latency}). In contrast, simply pausing for 200ms between consecutive forward passes largely mitigates power throttling, yielding more stable latency measurements (Appendix \ref{appendix:throtting}). Lastly, we find that prior work often reports latency using FP16 quantized models, but evaluates accuracy with FP32 models. However, naive quantization can significantly degrade performance (in some cases dropping performance to near 0 AP). To ensure fair comparison, we advocate for reporting accuracy and latency with the same model artifact.  We release our stand-alone benchmarking tool on \href{https://github.com/roboflow/single_artifact_benchmarking}{GitHub}. 

{
\setlength{\tabcolsep}{0.65em}
\begin{table}[t]
\caption{\textbf{COCO Detection Evaluation.} We compare \method \ with popular real-time and open-vocabulary object detectors below. We find that \method \ (nano) outperforms D-FINE (nano) and LW-DETR (tiny) by more than 5 AP. \method \ significantly outperforms YOLOv8 and YOLOv11, while \method's nano size achieves performance parity with YOLOv8 and YOLOv11's medium size model. We denote models that do not support TensorRT execution with a star, and instead report PyTorch latency results. See Appendix \ref{appendix:scaling} for L, XL, and Max variants of \method \ on COCO.}
\label{tab:coco_det}
\scalebox{0.7}{
\begin{tabular}{|lcccccccccc|}
\hline
\rowcolor{gray!10}
\textbf{Model} & \multicolumn{1}{c|}{\textbf{Size}} & \textbf{\# Params.} & \textbf{GFLOPS} & \multicolumn{1}{c|}{\textbf{Latency (ms)}} & \textbf{AP} & \textbf{AP$_{50}$} & \textbf{AP$_{75}$} & \textbf{AP$_S$} & \textbf{AP$_M$} & \textbf{AP$_L$} \\ \hline
\rowcolor{gray!10}
\multicolumn{11}{|l|}{\textbf{Real-Time Object Detection w/ NMS}}  \\ \hline
\multicolumn{1}{|l|}{YOLOv8 \citep{yolov8}}     & \multicolumn{1}{c|}{N}     &    3.2M   &    8.7    & \multicolumn{1}{c|}{2.1}             &  35.2  &   49.2   &   38.3   &  15.8   &  38.8   &  51.3   \\ \hline
\multicolumn{1}{|l|}{YOLOv11 \citep{yolov11}}     & \multicolumn{1}{c|}{N}     &   2.6M    &    6.5    & \multicolumn{1}{c|}{2.2}             &  37.1  &   51.6   &  40.4    &  17.3   &  40.7   &  55.6   \\ \hline \hline
\multicolumn{1}{|l|}{YOLOv8 \citep{yolov8}}     & \multicolumn{1}{c|}{S}     &    11.2M   &    28.6    & \multicolumn{1}{c|}{2.9}             &  42.4  &  57.6    &   46.0   &  22.2   &   47.1  &  59.6   \\ \hline
\multicolumn{1}{|l|}{YOLOv11 \citep{yolov11}}     & \multicolumn{1}{c|}{S}     &    9.4M   &    21.5    & \multicolumn{1}{c|}{3.2}             &  44.1  &   59.3   &   47.9   &  26.1   &  48.5   &  62.6   \\ \hline \hline
\multicolumn{1}{|l|}{YOLOv8 \citep{yolov8}}     & \multicolumn{1}{c|}{M}     &    25.9M   &    78.9    & \multicolumn{1}{c|}{5.4}             &  47.3  &   62.5   &   51.5   &  27.5   &  52.9   &   65.1  \\ \hline
\multicolumn{1}{|l|}{YOLOv11 \citep{yolov11}}     & \multicolumn{1}{c|}{M}     &    20.1M   &    68.0    & \multicolumn{1}{c|}{5.1}             &  48.3  &   63.6   &  52.5    &   29.1  &   53.8  &  66.3   \\ \hline
\rowcolor{gray!10}
\multicolumn{11}{|l}{\textbf{Open-Vocabulary Object Detection (Fully-Supervised Fine-Tuning)}}                                                                \\ \hline
\multicolumn{1}{|l|}{GroundingDINO \citep{liu2023grounding}}   & \multicolumn{1} {c|}{T}     &   173.0M   &   1008.3     & \multicolumn{1}{c|}{427.6*}             &  58.2  &   -   &  -   &  -   &  -   &   -  \\ \hline
\rowcolor{gray!10}
\multicolumn{11}{|l|}{\textbf{End-to-End Real-Time Object Detection}}                                                                \\ \hline
\multicolumn{1}{|l|}{LW-DETR \citep{chen2024lw} }      & \multicolumn{1}{c|}{T}     &   12.1M    &    21.4    & \multicolumn{1}{c|}{1.9}             &  42.9  &   60.7   &   45.9   &  22.7   &   47.3  &  60.0   \\ \hline
\multicolumn{1}{|l|}{D-FINE \citep{peng2024dfine}}      & \multicolumn{1}{c|}{N}     &    3.8M   &   7.3     & \multicolumn{1}{c|}{2.1}             &  42.7  &   60.2   &   45.4   &  22.9   &  46.6   &  62.1   \\ \hline
\multicolumn{1}{|l|}{\method \ (Ours)}     & \multicolumn{1}{c|}{N}     &   30.5M    &   31.9    & \multicolumn{1}{c|}{2.3}             &  48.0  &   67.0   &   51.4   &  25.2   &   53.5  &   70.0  \\ \hline \hline
\multicolumn{1}{|l|}{LW-DETR \citep{chen2024lw} }      & \multicolumn{1}{c|}{S}     &   14.6M    &    31.8    & \multicolumn{1}{c|}{2.6}             &  48.0  &   66.8   &   51.6   &  26.7   &   52.5  &  65.6   \\ \hline
\multicolumn{1}{|l|}{D-FINE \citep{peng2024dfine}}      & \multicolumn{1}{c|}{S}     &   10.2M    &   25.2     & \multicolumn{1}{c|}{3.5}             &  50.6  &  67.6    &  55.0    &  32.6   &  54.6   &  66.6   \\ \hline
\multicolumn{1}{|l|}{\method \ (Ours)}     & \multicolumn{1}{c|}{S}     &    32.1M   &   59.8     & \multicolumn{1}{c|}{3.5}             &  52.9  &  71.9    &   57.0   &  32.0   &  58.3   &  73.0   \\ \hline \hline
\multicolumn{1}{|l|}{RT-DETR \citep{zhao2024rtdetr}}      & \multicolumn{1}{c|}{R18}     &   36.0M    &    100.0    & \multicolumn{1}{c|}{4.4}             &  49.0  &   66.6   &   53.3   &  32.8   &   52.1  &   65.0  \\ \hline
\multicolumn{1}{|l|}{LW-DETR \citep{chen2024lw}}      & \multicolumn{1}{c|}{M}     &    28.2M   &    83.9    & \multicolumn{1}{c|}{4.4}             &  52.6  &   72.0   &   56.6   &   32.5  &  57.6   &  70.5   \\ \hline
\multicolumn{1}{|l|}{D-FINE \citep{peng2024dfine}}      & \multicolumn{1}{c|}{M}     &    19.2M   &    56.6    & \multicolumn{1}{c|}{5.4}             &  55.0  &   72.6   &   59.7   &  37.6   &  59.4   &  71.7   \\ \hline
\multicolumn{1}{|l|}{\method \ (Ours)}     & \multicolumn{1}{c|}{M}     &  33.7M     &   78.8     & \multicolumn{1}{c|}{4.4}             &  54.7  &   73.5   &   59.2   &  36.1   &  59.7   &  73.8   \\ \hline \hline
\multicolumn{1}{|l|}{\method \ (Ours)}     & \multicolumn{1}{c|}{2XL}     &    126.9M   &    438.4    & \multicolumn{1}{c|}{17.2}             &  60.1   &   78.5   &   65.5   &   43.2   &   64.9   &   76.2\\ \hline
\end{tabular}
}
\end{table}
}

\textbf{Evaluating \method \ and \method-Seg on COCO.} COCO \citep{lin2014coco} is a flagship benchmark for object detection and instance segmentation. In Table \ref{tab:coco_det}, we compare \method \ with leading real-time and open-vocabulary detectors. \method \ (nano) beats both D-FINE (nano) and LW-DETR (nano) by more than 5 AP. We see similar trends for small and medium sizes as well. Notably, \method \ also significantly outperforms YOLOv8 and YOLOv11. \method \ (nano) matches the performance of YOLOv8 and YOLOv11 (medium). We use mmdetection's implementation of GroundingDINO and include their reported AP since they do not release a model artifact for GroundingDINO fine-tuned on COCO. We benchmark mmGroundingDINO's parameter count, GFLOPS, and latency using the released open-vocabulary model. In Table \ref{tab:coco_seg}, we compare \method-Seg with real-time instance segmentation models. \method-Seg \ (nano) outperforms YOLOv8 and YOLOv11 at all sizes. Furthermore, \method-Seg (nano) beats FastInst by 5.4\% while running almost ten times faster. Similarly, \method \ (x-large) surpasses GroundingDINO (tiny), and \method-Seg (large) outperforms MaskDINO (R50), at a fraction of their runtime. 

{
\setlength{\tabcolsep}{0.7em}
\begin{table}[t]
\caption{\textbf{COCO Instance Segmentation Evaluation.} We compare \method \ with popular real-time instance segmentation methods on COCO. Notably, \method \ (nano) outperforms all reported YOLOv8 and YOLOv11 model sizes. Further \method \ (nano) outperforms FastInst by 5.4\%, while running nearly ten times faster. \method \ (medium) approaches the performance on MaskDINO at a fraction of the runtime. We denote models that do not support TensorRT execution with a star, and instead report PyTorch latency results.  Our latencies for YOLOs also include the conversion of protos into masks, which are not typically included in prior benchmarks but nonetheless contribute meaningfully to practical latency. See Appendix \ref{appendix:scaling} for L, XL, and Max variants of \method-Seg on COCO.
} 
\label{tab:coco_seg}
\scalebox{0.7}{
\begin{tabular}{|lcccccccccc|}
\hline
\rowcolor{gray!10}
\textbf{Model} & \multicolumn{1}{c|}{\textbf{Size}} & \textbf{\# Params.} & \textbf{GFLOPS} & \multicolumn{1}{l|}{\textbf{Latency (ms)}} & \textbf{AP} & \textbf{AP$_{50}$} & \textbf{AP$_{75}$} & \textbf{AP$_S$} & \textbf{AP$_M$} & \textbf{AP$_L$} \\ \hline
\rowcolor{gray!10}
\multicolumn{11}{|l|}{\textbf{Real-Time Instance Segmentation w/ NMS}}  \\ \hline
\multicolumn{1}{|l|}{YOLOv8 \citep{yolov8}}     & \multicolumn{1}{c|}{N}     &   3.4M    &    12.6    & \multicolumn{1}{c|}{3.5}             &  28.3  &   45.6   &  29.8    &   9.3  &   31.3  &  44.3   \\ \hline
\multicolumn{1}{|l|}{YOLOv11 \citep{yolov11}}     & \multicolumn{1}{c|}{N}     &   2.9M    &   10.4     & \multicolumn{1}{c|}{3.6}             &  30.0  &   47.8   &   31.5   &  10.0   &  33.4   &  47.7   \\ \hline \hline
\multicolumn{1}{|l|}{YOLOv8 \citep{yolov8}}     & \multicolumn{1}{c|}{S}     &   11.8M    &   42.6     & \multicolumn{1}{c|}{4.2}             &  34.0  &   53.8   &   36.0   &  13.6   &  38.5   &  52.2   \\ \hline
\multicolumn{1}{|l|}{YOLOv11 \citep{yolov11}}     & \multicolumn{1}{c|}{S}     &   10.1M    &   35.5     & \multicolumn{1}{c|}{4.6}             &  35.0  &   55.4   &   37.1   &   15.3  &  39.7   &   53.9  \\ \hline \hline
\multicolumn{1}{|l|}{YOLOv8 \citep{yolov8}}     & \multicolumn{1}{c|}{M}     &   27.3M    &   110.2     & \multicolumn{1}{c|}{7.0}             &  37.3  &  58.2    &  39.9    &  16.7   &  43.0   &  56.1   \\ \hline
\multicolumn{1}{|l|}{YOLOv11 \citep{yolov11}}     & \multicolumn{1}{c|}{M}     &   22.4M    &   123.3     & \multicolumn{1}{c|}{6.9}             &  38.5  &   60.0   &  40.9    &  18.0   &  44.3   &  57.6   \\ \hline
\rowcolor{gray!10}
\multicolumn{11}{|l|}{\textbf{End-to-End Instance Segmentation}}                                                                \\ \hline
\multicolumn{1}{|l|}{\method-Seg. (Ours)}     & \multicolumn{1}{c|}{N}     &   33.6M    &    50.0    & \multicolumn{1}{c|}{3.4}             &  40.3  &  63.0    &  42.6    &  16.3   &  45.3   &   63.6  \\ \hline \hline

\multicolumn{1}{|l|}{\method-Seg. (Ours)}     & \multicolumn{1}{c|}{S}     &   33.7M   &    70.6    & \multicolumn{1}{c|}{4.4}             &  43.1  &   66.2   & 45.9     &  21.9   &  48.5   &   64.1  \\ \hline \hline
\multicolumn{1}{|l|}{FastInst \citep{he2023fastinst}}      & \multicolumn{1}{c|}{R50}     &    29.7M   &   99.7     & \multicolumn{1}{c|}{39.6$^*$}             &  34.9  &  56.0    &  36.2    &  13.3   &  38.0   &   56.8  \\ \hline

\multicolumn{1}{|l|}{MaskDINO \citep{li2023maskdino}}      & \multicolumn{1}{c|}{R50}     &   52.1M    &   586     & \multicolumn{1}{c|}{242$^*$}             &  46.3  &   69.0   &   50.7   &  26.1   &  49.3   &   66.1  \\ \hline
\multicolumn{1}{|l|}{\method-Seg. (Ours)}     & \multicolumn{1}{c|}{M}     &    35.7M      &    102.0    & \multicolumn{1}{c|}{5.9}             &  45.3  &   68.4   &  48.8    &   25.5  &   50.4  &  65.3   \\ \hline \hline
\multicolumn{1}{|l|}{\method \ (Ours)}     & \multicolumn{1}{c|}{2XL}     &   38.6M    &    435.3    & \multicolumn{1}{c|}{21.8}             &   49.9  &   73.1   &   54.5   &   33.9   &   54.1   &   65.7  \\ \hline

\end{tabular}
}
\end{table}
}

\textbf{Evaluating \method \ on RF100-VL.} RF100-VL is a challenging detection benchmark composed of 100 diverse datasets. We report  latencies, FLOPs, and accuracy averaged over all 100 datasets in Table \ref{tab:rf100-vl}. Our results show that \method \ (2x-large) outperforms GroundingDINO and LLMDet while requiring only a fraction of their runtime. Interestingly, RT-DETR outperforms D-FINE (which is built on RT-DETR) at AP$_{50}$, indicating that D-FINE's hyperparameters are potentially overfit to COCO. 
We also note that \method \ benefits from scaling to larger backbone sizes (Appendix \ref{appendix:scaling}). In contrast, YOLOv8 and YOLOv11 consistently underperform DETR-based detectors, and scaling these model families to larger sizes does not improve their performance on RF100-VL (Figure \ref{fig:pareto-curve}). 

{
\setlength{\tabcolsep}{0.65em}
\begin{table}[t]
\caption{\textbf{RF100-VL Evaluation.} We compare \method \ with real-time and open-vocabulary object detectors on RF100-VL. Interestingly, \method \ (2x-large) outperforms GroundingDINO (tiny), and LLMDet (tiny) at a fraction of their runtime. We report the average latency and FLOPs over all 100 datasets. We note that YOLOv8 and YOLOv11's latency measurements may be suboptimal because the default tuned NMS threshold of 0.01 may not work well for all datasets in RF100-VL. We denote models that do not support TensorRT execution with a star, and instead report PyTorch latency results. See Appendix \ref{appendix:scaling} for L, XL, and Max variants of \method \ on RF100-VL.} 
\label{tab:rf100-vl}
\scalebox{0.7}{
\begin{tabular}{|lcccccccccc|}
\hline
\rowcolor{gray!10}
\textbf{Model} & \multicolumn{1}{c|}{\textbf{Size}} & \textbf{\# Params.} & \textbf{GFLOPS} & \multicolumn{1}{l|}{\textbf{Latency (ms)}} & \textbf{AP} & \textbf{AP$_{50}$} & \textbf{AP$_{75}$} & \textbf{AP$_S$} & \textbf{AP$_M$} & \textbf{AP$_L$} \\ \hline
\rowcolor{gray!10}
\multicolumn{11}{|l|}{\textbf{Real-Time Object Detectors w/ NMS}}  \\ \hline
\multicolumn{1}{|l|}{YOLOv8 \citep{yolov8}}     & \multicolumn{1}{c|}{N}     &   3.2M    &    8.7    & \multicolumn{1}{c|}{2.6}             &  55.0  & 81.1     &  59.5    &   4.8  &   44.1  &   48.0  \\ \hline
\multicolumn{1}{|l|}{YOLOv11 \citep{yolov11}}     & \multicolumn{1}{c|}{N}     &   2.6M    &    6.5    & \multicolumn{1}{c|}{3.0}             &  55.5  &  81.3    &  60.3    &  4.7   &   44.4  &   49.2  \\ \hline \hline
\multicolumn{1}{|l|}{YOLOv8 \citep{yolov8}}     & \multicolumn{1}{c|}{S}     &   11.2M    &    28.6    & \multicolumn{1}{c|}{3.1}             &  56.3  &   82.0   &  60.9    &   6.1  &  45.6   &  48.6   \\ \hline
\multicolumn{1}{|l|}{YOLOv11 \citep{yolov11}}     & \multicolumn{1}{c|}{S}     &   9.4M    &    21.5      & \multicolumn{1}{c|}{3.3}             &  56.4  &  82.5    &   61.3   &   6.5  &  45.5   &  48.5   \\ \hline \hline
\multicolumn{1}{|l|}{YOLOv8 \citep{yolov8}}     & \multicolumn{1}{c|}{M}     &   25.9M    &    78.9    & \multicolumn{1}{c|}{5.4}             & 56.5   &   82.3   &   60.9   &   6.4  &  45.7   &  48.6   \\ \hline
\multicolumn{1}{|l|}{YOLOv11 \citep{yolov11}}     & \multicolumn{1}{c|}{M}   &   20.1M    &    68.0         & \multicolumn{1}{c|}{5.1}             &  57.0  &   82.5   &   61.9   &   7.3  &  46.1   &  48.6   \\ \hline
\rowcolor{gray!10}
\multicolumn{11}{|l|}{\textbf{Open-Vocabulary Object-Detectors (Fully-Supervised Fine-Tuning)}}                                                                \\ \hline
\multicolumn{1}{|l|}{GroundingDINO \citep{liu2023grounding}}   & \multicolumn{1} {c|}{T}     &   173.0M   &   1008.3     & \multicolumn{1}{c|}{309.9$^{*}$ }             &  62.3  &   88.8   &   67.8   &  39.2   &  57.7   &   69.5  \\ \hline
\multicolumn{1}{|l|}{LLMDet \citep{fu2025llmdet}}     & \multicolumn{1}{c|}{T}     &   173.0M    &   1008.3     & \multicolumn{1}{c|}{308.4$^{*}$}             &  62.3  &   88.3   &   67.8   &  39.1   &  57.6   &   70.3  \\ \hline
\rowcolor{gray!10}
\multicolumn{11}{|l|}{\textbf{End-to-End Real-Time Object Detectors}}                                                                \\ \hline
\multicolumn{1}{|l|}{LW-DETR \citep{chen2024lw} }      & \multicolumn{1}{c|}{N}     &   12.1M   &   21.4    & \multicolumn{1}{c|}{1.9}             &  57.1  &  84.7   &  61.5   &  31.2   &  51.8  & 65.8   \\ \hline
\multicolumn{1}{|l|}{D-FINE \citep{peng2024dfine}}      & \multicolumn{1}{c|}{N}     &   3.8M    &   7.3     & \multicolumn{1}{c|}{2.0}             &  58.2  &  84.4    &   62.5   &  32.4   &  52.9   &  65.8   \\ \hline
\multicolumn{1}{|l|}{\method \ (Ours)}     & \multicolumn{1}{c|}{N}     &   31.2M   &   34.5     & \multicolumn{1}{c|}{2.5}             & 57.8   &   85.1   &   62.5   &  30.1   &   52.2   &   67.2  \\ \hline
\multicolumn{1}{|l|}{\method \ w/ Fine-Tuning (Ours)}     & \multicolumn{1}{c|}{N}     &   31.2M   &   34.5     & \multicolumn{1}{c|}{2.5}             &  58.6  &  85.7   &   63.0  &  31.0  &  53.2   &  67.6  \\ \hline \hline
\multicolumn{1}{|l|}{LW-DETR \citep{chen2024lw} }      & \multicolumn{1}{c|}{S}     &   14.6M   &    31.8     & \multicolumn{1}{c|}{2.6}             &  57.4  &   85.0   &   62.0   &  32.1   &  52.1   &  65.8   \\ \hline
\multicolumn{1}{|l|}{D-FINE \citep{peng2024dfine}}      & \multicolumn{1}{c|}{S}  &  10.2M    &   25.2         & \multicolumn{1}{c|}{3.5}             &  60.3  &  85.3    &   65.4   &  36.6   &  56.0   &  68.4   \\ \hline
\multicolumn{1}{|l|}{\method \ (Ours)}     & \multicolumn{1}{c|}{S}     &    33.5M   &    62.4    & \multicolumn{1}{c|}{3.7}             & 60.9   &   87.5   &   66.1   &  34.2   &   55.7  &  69.6   \\ \hline
\multicolumn{1}{|l|}{\method \ w/ Fine-Tuning (Ours)}     & \multicolumn{1}{c|}{S}     &    33.5M   &    62.4    & \multicolumn{1}{c|}{3.7}             &  61.2  &  87.7    &  66.1   &  34.9  &  55.6  & 69.5   \\ \hline \hline
\multicolumn{1}{|l|}{RT-DETR \citep{zhao2024rtdetr}}      & \multicolumn{1}{c|}{M}     &   36.0M    &    100.0    & \multicolumn{1}{c|}{4.3}             & 59.6   &   85.7   &   64.6   &  36.4   &  54.6   &  67.3   \\ \hline
\multicolumn{1}{|l|}{LW-DETR \citep{chen2024lw}}      & \multicolumn{1}{c|}{M}     &   28.2M    &   83.9     & \multicolumn{1}{c|}{4.3}             & 59.8   &   86.8   &   64.9   &  34.0   &  54.4   &   68.9  \\ \hline
\multicolumn{1}{|l|}{D-FINE \citep{peng2024dfine}}      & \multicolumn{1}{c|}{M}     &   19.2M    &    56.6    & \multicolumn{1}{c|}{5.6}             &   60.6 &   85.5   &   65.8   &  36.0   &  56.6   &  67.5   \\ \hline
\multicolumn{1}{|l|}{\method \ (Ours)}     & \multicolumn{1}{c|}{M}     &   33.5M    &   86.7     & \multicolumn{1}{c|}{4.6}             & 61.7   &    88.0  &   66.9   &  35.8   &   56.5  &   70.0  \\ \hline 
\multicolumn{1}{|l|}{\method \ w/ Fine-Tuning (Ours)}     & \multicolumn{1}{c|}{M}     &   33.5M    &   86.7     & \multicolumn{1}{c|}{4.6}             & 62.0   &  88.1   &  67.1   &  36.2   &  56.4  &  70.2   \\ \hline
\multicolumn{1}{|l|}{\method \ (Ours)}     & \multicolumn{1}{c|}{2XL}     &  123.5M   &   410.2    & \multicolumn{1}{c|}{15.6}             &  63.3   &   88.9   &    69.0  &  38.7    &   58.2   &   71.6   \\ \hline
\multicolumn{1}{|l|}{\method \ (Ours) w/ Fine-Tuning}     & \multicolumn{1}{c|}{2XL}     &   123.5M   &   410.2     & \multicolumn{1}{c|}{15.6}             &  63.5  &   89.0   &  69.2    &  38.9   &  58.3   &  71.7   \\ \hline
\end{tabular}
}
\end{table}
}

\textbf{Impact of Neural Architecture Search.} We ablate the impact of weight-sharing NAS in Table \ref{fig:nas-knobs}. We find that adopting a gentler set of hyperparameters compared to LW-DETR (e.g. larger batch size, lower learning rate, and replacing batch normalization with layer normalization) reduces performance over LW-DETR by 1.0\%. Notably, replacing batch normalization with layer normalization hurts performance, but is necessary to train on consumer hardware. However, replacing LW-DETR's CAEv2 backbone with DINOv2 improves performance by 2\%. The lower learning rate, in particular, helps preserve DINOv2's pre-trained knowledge, while additional epochs of Objects-365 pre-training further compensate for the slower optimization. Our final model with weight-sharing NAS improves over LW-DETR by 2\% without increasing latency.

{
\setlength{\tabcolsep}{0.8em}
\begin{table}[t]
\caption{\textbf{Ablation on Neural Architecture Search.} We ablate the impact of each ``tunable knob'' on accuracy and latency below. Using a gentler set of hyperparameters compared to LW-DETR (e.g. smaller batch size, lower learning rate, replacing batch norm with layer norm) reduces performance by 1\%. However, we regain this lost performance by replacing LW-DETR's CAEV2 backbone with DINOv2. Importantly, the lower learning rate and layer-norm allow us to better preserve DINOv2's foundational knowledge and allows us to train with larger batch sizes, making weight-sharing NAS more effective. Counterintuitively, introducing weight sharing NAS to the training scheme improves the performance of the base configuration even though patch size 14 isn't in the NAS search space.} 
\label{tab:nas}
\scalebox{0.7}{
\begin{tabular}{|lccccccccc|}
\hline
\rowcolor{gray!10}
\multicolumn{1}{|l|}{\textbf{Model}} & \textbf{\# Params.} & \textbf{GFLOPS} & \multicolumn{1}{l|}{\textbf{Latency (ms)}} & \textbf{AP} & \textbf{AP$_{50}$} & \textbf{AP$_{75}$} & \textbf{AP$_S$} & \textbf{AP$_M$} & \textbf{AP$_L$} \\ \hline
\multicolumn{1}{|l|}{LW-DETR (M)}     &  28.2M   &   83.7   &    \multicolumn{1}{c|}{4.4}   & 52.6 & 72.0 & 56.6 &  32.5 & 57.6 & 70.5  \\ \hline
\multicolumn{1}{|l|}{+ Gentler Hyperparameters}     &      28.2M      &    83.7   & \multicolumn{1}{c|}{4.4 }             &  51.6 &   71.1   & 55.5 & 31.7 & 56.4 & 69.4  \\ \hline
\multicolumn{1}{|l|}{+ DINOv2 Backbone}     &       32.3M     &     78.2   &  \multicolumn{1}{c|}{4.7}             & 53.6 & 72.7  & 58.0 & 34.3 & 58.3 & 72.4 \\ \hline
\multicolumn{1}{|l|}{+ Additional O365 Pre-Training}     &     32.3M       &    78.2    &  \multicolumn{1}{c|}{4.7}             & 54.3 & 73.4 & 58.8 & 35.8 & 59.2 & 72.3 \\ \hline
\multicolumn{1}{|l|}{+ Weight Sharing NAS}    &     32.3M       &    78.2   &  \multicolumn{1}{c|}{4.7}   &  54.6 &      73.4  &  59.3    &  36.3    &  59.3   &  72.1    \\ \hline
\multicolumn{1}{|l|}{\quad + \small Patch Size 14$\,\to\,$16, Res 560$\,\to\,$640}     &   32.3M   & 78.5 &  \multicolumn{1}{c|}{4.7}             &  54.4  &  73.2    &   59.1   &  35.9   &  59.2   &  72.1   \\ \hline
\multicolumn{1}{|l|}{\quad + Image Resolution 640$\,\to\,$576}     &  32.2M  & 64.2 &    \multicolumn{1}{c|}{4.0}             &  53.6  &  72.4    &   58.2   &  34.8   &  58.6   &  72.0   \\ \hline
\multicolumn{1}{|l|}{\quad + \# Windows per Block 4$\,\to\,$2 }     &   32.2M       &  63.7    &  \multicolumn{1}{c|}{4.3}             &  54.3  &  73.3    & 58.8 & 35.6 & 59.4 & 73.2 \\ \hline
\multicolumn{1}{|l|}{\quad + \# Decoder Layers 3$\,\to\,$4}     &     33.7M       &   64.8   &  \multicolumn{1}{c|}{4.4}             &  54.6  & 73.5 & 59.1 & 36.0 & 59.8 & 73.7 \\ \hline
\multicolumn{1}{|l|}{\quad + \# Query Tokens 300$\,\to\,$300}    &    33.7M        &   64.8   & \multicolumn{1}{c|}{4.4}             &  54.6  & 73.5 & 59.1 & 36.0 & 59.8 & 73.7 \\ \hline
\end{tabular}
}
\end{table}
}

\textbf{Impact of Backbone Architecture and Pre-Training.} 
We study the impact of different backbone architectures in \method. We find that DINOv2 achieves the best performance, outperforming CAEv2 by 2\%. Interestingly, despite having fewer parameters than SigLIPv2, SAM2's Hiera-S backbone is considerably slower. This is in contrast with Hiera's claim that it is meaningfully faster than equivalently performant ViTs. However, Hiera does not explore latency in the context of Flash Attention kernels, which are highly optimized in compilers such as TensorRT. Additionally, existing foundation model families typically do not release lightweight ViT variants such as ViT-S or ViT-T, making it difficult to repurpose such models for real-time applications.

{
\setlength{\tabcolsep}{0.6em}
\begin{table}[t]
\caption{\textbf{Ablation on Backbone.} We ablate the impact of using different backbone architectures for \method \ below. We find that DINOv2 achieves the highest performance, outperforming CAEv2 by 2.4\%. All models are pretrained with 60 epochs of Objects-365 and the ``Gentler Hyperparameters'' setting. Note that SAM2 and SigLIPv2 perform poorly when evaluated in FP16. Therefore, we report FP16 TensorRT latency with FP32 ONNX accuracy for these two models as an upper bound on their performance if optimized for FP16.} 
\label{tab:backbone}
\scalebox{0.7}{
\begin{tabular}{|lccccccccc|}
\hline
\rowcolor{gray!10}
\multicolumn{1}{|l|}{\textbf{LW-DETR (M) + Gentler Hyperparameters}} & \textbf{\# Params.} & \textbf{GFLOPS} & \multicolumn{1}{l|}{\textbf{Latency (ms)}} & \textbf{AP} & \textbf{AP$_{50}$} & \textbf{AP$_{75}$} & \textbf{AP$_S$} & \textbf{AP$_M$} & \textbf{AP$_L$} \\ \hline
\multicolumn{1}{|l|}{\quad w/ CAEv2 ViT/S-16-Truncated Backbone}     &  28.3M   & 83.7 &  \multicolumn{1}{c|}{4.4}             &  52.3  &  71.4    &   56.3   &  32.3   &  56.4   &  70.0   \\ \hline
\multicolumn{1}{|l|}{\quad w/ DINOv2 ViT/S-14 Backbone}     &      32.3M      &   78.2     &  \multicolumn{1}{c|}{4.7}             &  54.3  &  73.4    &   58.8   &  35.8   & 59.2    &  72.3   \\ \hline
\multicolumn{1}{|l|}{\quad w/ SigLIPv2 ViT/B-32 Backbone$^*$} & 105.1M     &     81.6   &  \multicolumn{1}{c|}{4.8}             & 50.4   & 70.4  & 53.7 & 28.0 & 55.3 & 73.0 \\ \hline
\multicolumn{1}{|l|}{\quad w/ SAM2 Hiera-S Backbone$^*$} & 44.0M &  109.1 &  \multicolumn{1}{c|}{11.2}   &  53.6   &  72.4    &  57.9    &  33.3   &   58.3  &  71.0   \\ \hline
\end{tabular}
}
\end{table}
}

\textbf{Rethinking Standard Accuracy Benchmarking Practices.} Following prior work, we report all COCO results on the validation set. However, relying solely on the validation for both model selection and evaluation can lead to overfitting. For example, D-FINE (which builds on RT-DETR) conducts an extensive hyperparameter sweep on COCO's validation set and reports its best model. However, evaluating this configuration on RF100-VL shows that D-FINE underperforms RT-DETR on the test set. In contrast, our method achieves state-of-the-art performance among all real-time detectors on both RF100-VL and COCO, demonstrating the robustness of our weight-sharing NAS. In addition to evaluating on COCO, we advocate that future detectors should also evaluate on datasets with public validation and test splits like RF100-VL.

\textbf{Limitations.} Despite controlling for power throttling and GPU overheating during inference, our latency measurements still have a variance of up to 0.1ms due to the non-deterministic behavior of TensorRT during compilation. Specifically, TensorRT can introduce power throttling, which in turn affects the resulting engine and leads to random fluctuations in latency. Although the measurement of a given TensorRT engine is generally consistent, recompiling the same ONNX artifact can produce slightly different latency results. Therefore, we only report latencies with one digit of precision after the decimal place. 

\section{Conclusion}
In this paper, we introduce \method, a state-of-the-art NAS-based method for fine-tuning specialist end-to-end object detectors for diverse target datasets and hardware platforms. Our approach outperforms prior state-of-the-art real-time methods on COCO and RF100-VL, improving upon D-FINE (nano) by 5\% AP on COCO. Moreover, we highlight that current architectures, learning rate schedulers and augmentation schedulers are tailored to maximize performance on COCO, suggesting that the community should benchmark models on diverse, large-scale datasets to prevent implicit overfitting. Lastly, we highlight the high variance in latency benchmarking due to power throttling and propose a standardized protocol to improve reproducibility.

\section*{Acknowledgements}
This work was supported in part by compute provided by NVIDIA DGX. We'd like to thank Brad Dwyer for reviewing early versions of our manuscript. 

\clearpage
\newpage

\bibliography{iclr2026_conference}
\bibliographystyle{iclr2026_conference}

\newpage 

\appendix

\section{Implementation Details}
\label{sec:implementation_details}
\textbf{Training Hyperparamters.} \method \ extends LW-DETR \citep{chen2024lw} for Neural Architecture Search. We highlight key differences in our training procedure below. First, we pseudo-label Objects-365 \citep{objects365} with SAM2 \citep{ravi2024sam2} to allow us to pre-train the segmentation and detection heads on the same data. We use a learning rate of 1e-4 (LW-DETR uses 4e-4), and a batch size of 128 (LW-DETR uses the same). Similar to DINOv3 \citep{simeoni2025dinov3}, we use an EMA scheduler since this is necessary for EMA to function. However, unlike DINOv3, we omit learning-rate warm-up. We clip all gradients greater than 0.1 and apply a per-layer multiplicative decay of 0.8 to preserve information (especially the earlier layers) in the DINOv2 backbone. We place our window attention blocks between layers \{0, 1, 3, 4, 6, 7, 9, 10\}, while LW-DETR places their window attention blocks between layers \{0, 1, 3, 6, 7, 9\}. Although we have the same number of windows, contiguous windowed blocks don't require an additional reshape operation, making our implementation slightly more efficient. Further, we train with more multi-scale resolutions (0.5 to 1.5 scale) than LW-DETR (0.7 to 1.4 scale) to ensure that the augmentation is symmetric around the default scale. Notably, we add resolution as a ``tunable knob'' in our NAS search space, while LW-DETR uses it as a form of data augmentation. Our model training and inference code is available on \href{https://github.com/roboflow/rf-detr}{GitHub}.

\textbf{Latency Evaluation.} We ensure fair evaluation between models by measuring detection accuracy and latency using the same artifact. To further standardize inference, we employ CUDA graphs in TensorRT, which pre-queue all kernels rather than requiring the CPU to launch them serially during execution. This optimization can accelerate some networks depending on the number and type of kernels used by the model. We observe that RT-DETR, LW-DETR, and \method \ benefit from this optimization. Further, CUDA graphs place LW-DETR on the same latency-accuracy curve as D-FINE, since CUDA graphs speed up LW-DETR but do not benefit D-FINE. We release our stand-alone latency benchmarking tool on \href{https://github.com/roboflow/single_artifact_benchmarking}{GitHub}.

\textbf{Pareto-Optimal Model Configurations in COCO.} We present the Pareto-Optimal \method \ and \method-Seg configurations in Tables \ref{tab:coco_det_config} and \ref{tab:coco_seg_config}. We highlight notable trends about \method's Pareto-Optimal architectures in Appendix \ref{appendix:trends}.

{
\setlength{\tabcolsep}{1.0em}
\begin{table}[h]
\caption{\textbf{\method \ COCO Detection Model Config.}}
\label{tab:coco_det_config}
\scalebox{0.8}{
\begin{tabular}{|c|c|c|c|c|c|c|}
\hline
\rowcolor{gray!10}
\textbf{Model Size} & \textbf{Resolution} & \textbf{Patch Size} & \textbf{Windows} & \textbf{Decoder Layers} & \textbf{Queries} & \textbf{Backbone} \\ \hline
N         &          384           &          16           &       2           &          2               &          300        &          DINOv2-S            \\ \hline
S         &          512           &         16            &        2          &          3               &         300         &            DINOv2-S          \\ \hline
M         &          576           &          16           &          2        &           4              &        300          &          DINOv2-S            \\ \hline
L         &          704           &         16            &         2         &            4             &       300           &          DINOv2-S            \\ \hline
XL        &          700           &           20          &        1          &           5              &       300           &          DINOv2-B            \\ \hline
2XL       &            880         &           20          &         2         &            5             &         300         &          DINOv2-B            \\ \hline
Max       &         828            &         12            &       1           &            6             &          300        &          DINOv2-B            \\ \hline
\end{tabular}
}
\end{table}
}

{
\setlength{\tabcolsep}{1.0em}
\begin{table}[h]
\caption{\textbf{\method-Seg COCO Segmentation Model Config.}}
\label{tab:coco_seg_config}
\scalebox{0.8}{
\begin{tabular}{|c|c|c|c|c|c|c|}
\hline
\rowcolor{gray!10}
\textbf{Model Size} & \textbf{Resolution} & \textbf{Patch Size} & \textbf{Windows} & \textbf{Decoder Layers} & \textbf{Queries} & \textbf{Backbone} \\ \hline
N         &         312            &         12            &      1            &             4            &          100        &           DINOv2-S           \\ \hline
S         &          384           &         12            &         2         &               4          &          100        &          DINOv2-S            \\ \hline
M         &          432           &        12             &       2           &             5            &        200          &        DINOv2-S              \\ \hline
L         &        504             &          12           &       2           &                5         &           300       &            DINOv2-S          \\ \hline
XL        &          624           &         12            &         2         &               6          &        300          &         DINOv2-S             \\ \hline
2XL       &         768            &           12          &      2            &              6           &          300        &          DINOv2-S            \\ \hline
Max       &          890           &          10           &         1         &             6           &         300         &           DINOv2-S           \\ \hline
\end{tabular}
}
\end{table}
}

\textbf{Parameter Sampling Grid.} Lastly, we present our sampling grid for training and inference below. Importantly, we only drop decoder layers and queries during inference. We uniformly sample configurations during training and perform grid search over all configurations during inference to find Pareto-Optimal model configurations. RF-DETR’s total training time is roughly two to four times as long as a non-NAS baseline, depending on the target dataset. However, \method \ can generate all size configurations from this single training run, while other non-NAS baselines must be re-trained for each new model size. We evaluate 6,468 network configurations (11 resolutions * 7 patch sizes * 7 decoder layers * 3 windows * 4 query settings) during architecture search. We estimate that this search takes approximately 10,000 GPU Hours (200 T4 GPUs * 48 hours). 

\textbf{Training Configurations}
\begin{itemize}
    \item Image Resolutions: 320, 384, 448, 512, 576, 640, 704, 768, 832, 896, 960
    \item Patch Sizes: 8, 10, 12, 16, 20, 24, 32
    \item Number of Decoder Layers: 6
    \item Number of Windows: 1, 2, 4
    \item Number of Queries: 300
\end{itemize}

\textbf{Inference Configurations}
\begin{itemize}
    \item Image Resolutions: 320, 384, 448, 512, 576, 640, 704, 768, 832, 896, 960
    \item Patch Sizes: 8, 10, 12 ,14, 16, 20, 24, 32
    \item Number of Decoder Layers: 0, 1, 2, 3, 4, 5, 6
    \item Number of Windows: 1, 2, 4
    \item Number of Queries: 50, 100, 200, 300
\end{itemize}

\section{Ablation on Query Tokens and Decoder Layers}
\label{appendix:query_decoder}

\begin{figure}[h]
    \centering
    \includegraphics[width=0.98\linewidth]{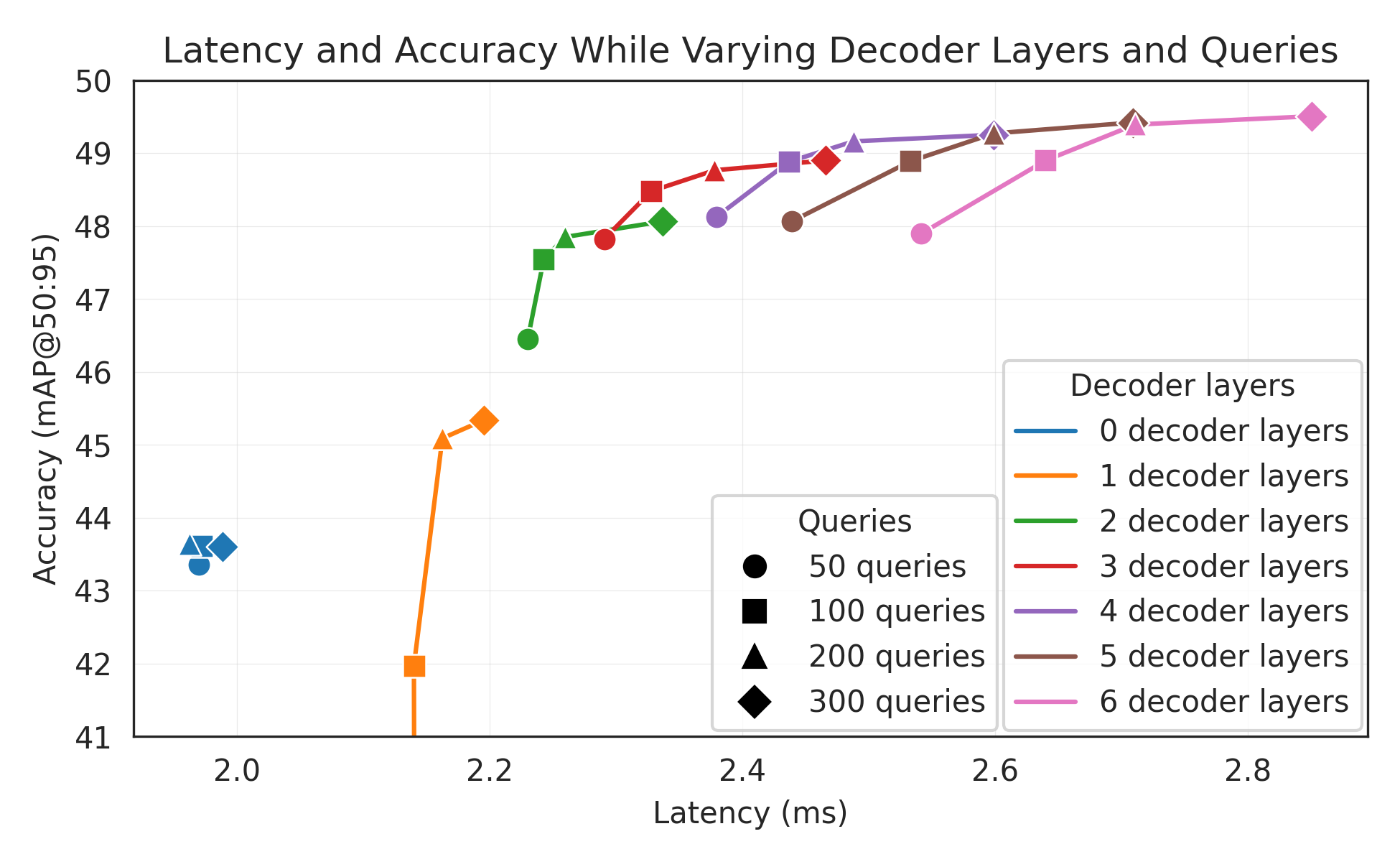}
    \caption{{\bf Impact of Decoder Layers vs. Query Tokens}. We evaluate the impact of inference-time query dropping for trading-off accuracy and latency in \method \ (nano). Interestingly, we find that dropping the 100 lowest confidence queries does not significantly reduce performance, but modestly improves latency for all decoder layers.}
    \label{fig:nas_query}
\end{figure}

We train \method \ (nano) with 300 object queries, following standard practices for real-time DETR-based object detectors. However, many datasets contain fewer than 300 objects per image. Therefore, processing all 300 queries can be computationally wasteful. LW-DETR (tiny) demonstrates that training with fewer queries can improve the latency-accuracy tradeoff. Rather than deciding on the optimal number of queries apriori, we find that we can drop queries at test time \textit{without retraining} by discarding the lowest-confidence queries ordered by the confidence of the corresponding token at the output of the encoder. As shown in Figure \ref{fig:nas_query}, this yields meaningful latency-accuracy tradeoffs. In addition, prior work \citep{zhao2024rtdetr} demonstrates that decoder layers can be pruned at test time, since each layer is supervised independently during training. We find that it is possible to remove \emph{all} decoder layers, relying solely on the initial query proposals from the two-stage DETR pipeline. In this case, there is no cross-attention to the encoder states or self-attention between queries, leading to a substantial runtime reduction. The resulting model resembles a single-stage YOLO-style architecture without NMS. Eliminating the final decoder layer reduces latency by 10\% with only a 2 mAP drop in performance.

\section{Benchmarking FLOPs}
We benchmark FLOPs for \method, GroundingDINO, and LLMDet with PyTorch's \href{https://github.com/pytorch/pytorch/blob/baee623691a38433d10843d5bb9bc0ef6a0feeef/torch/utils/flop_counter.py#L596}{\texttt{FlopCounterMode}}. We find that \texttt{FlopCounterMode} closely reproduces FLOP counts obtained with custom benchmarking tools for YOLOv11, D-FINE, and LW-DETR. In practice, we also find that it provides more reliable results than CalFLOPs \citep{calflops}. Notably, LW-DETR's FLOPs count is roughly twice that of the originally reported result (cf. Table \ref{tab:flops}). We posit that this discrepancy can be attributed to LW-DETR reporting MACs instead of FLOPs. We rely on the officially reported FLOPs counts from YOLOv11, YOLOv8, D-FINE, and RT-DETR.

{
\setlength{\tabcolsep}{3.4em}
\begin{table}[h]
\caption{\textbf{FLOPs Benchmarking Comparison.} We compare FLOPs reported with custom benchmarking tools, CalFLOPs, and PyTorch's FlopCounterMode. Notably, we find that FlopCounterModel closely matches the results reported with custom benchmarking code, suggesting that it is more reliable than prior generic benchmarking tools.}
\label{tab:flops}
\scalebox{0.7}{
\begin{tabular}{|lc|ccc|}\hline
\rowcolor{gray!10}
\hline
\textbf{Model} & \textbf{Size} & \textbf{Reported} & \textbf{CalFLOPs} & \textbf{FlopCounterMode} \\
\hline
D-FINE & S  & 25.2 M & 25.2 M & 25.5 M \\ \hline
LW-DETR & S & 16.6 M & 22.9 M & 31.8 M \\ \hline 
YOLO11 & S  & 21.5 M & 23.9 M & 21.6 M \\ \hline
\end{tabular}
}
\end{table}
}

\section{Impact of Class-Names on Open-Vocabulary Detectors}
We evaluate the impact of fine-tuning open-vocabulary detectors like GroundingDINO with class names on RF100-VL in Table \ref{tab:rf100-vl-class-names}. Intuitively, GroundingDINO's vision-language pre-training should be more more useful when we prompt with class names (e.g. {\tt car}, {\tt truck}, {\tt bus}) instead of class indices (e.g. 0, 1, 2). Using class names when fine-tuning provides more information to the VLM about the underlying data than is available to non-VLM detectors, potentially leading to better downstream performance. However, we find that fine-tuning GroundingDINO on RF100-VL yields nearly identical performance in both cases, suggesting that naively fine-tuning the end-to-end model mitigates the benefits of open-vocabulary pre-training. Future work should investigate ways of effectively fine-tuning VLMs to preserve foundational pre-training.

{
\setlength{\tabcolsep}{0.2em}
\begin{table}[h]
\caption{\textbf{Evaluating the Impact of Class Names.} We evaluate the impact of using class-names when fine-tuning VLMs like GroundingDINO. We find that class-names do not provide significant benefit over prompting with class indices, suggesting that fine-tuning has diminished the impact of internet-scale pre-training.
} 
\label{tab:rf100-vl-class-names}
\scalebox{0.7}{
\begin{tabular}{|lcccccccccc|}
\hline
\rowcolor{gray!10}
\textbf{Model} & \multicolumn{1}{c|}{\textbf{Size}} & \textbf{\# Params.} & \textbf{GFLOPS} & \multicolumn{1}{l|}{\textbf{Latency (ms)}} & \textbf{AP} & \textbf{AP$_{50}$} & \textbf{AP$_{75}$} & \textbf{AP$_S$} & \textbf{AP$_M$} & \textbf{AP$_L$} \\ \hline
\rowcolor{gray!10}
\multicolumn{11}{|l|}{\textbf{RF100-VL}}                                                                \\ \hline
\multicolumn{1}{|l|}{GroundingDINO \citep{liu2023grounding} w/ Standard Class Names}   & \multicolumn{1}{c|}{T}     &   173.0M   &   1008.3     & \multicolumn{1}{c|}{309.9$^{*}$ }             &  62.3  &   88.8   &   67.8   &  39.2   &  57.7   &   69.5  \\ \hline
\multicolumn{1}{|l|}{GroundingDINO \citep{liu2023grounding} w/ Class Index Names}   & \multicolumn{1}{c|}{T}     &   173.0M   &   1008.3     & \multicolumn{1}{c|}{309.9$^{*}$ }             &  62.5  &   88.2   &   68.3   &  40.0   &   58.4  &  70.3   \\\hline
\end{tabular}
}
\end{table}
}

\section{Benchmarking Larger Model Variants}
\label{appendix:scaling}

Detectors like LW-DETR \citep{chen2024lw} and D-FINE \citep{peng2024dfine} hand-design larger variants to scale up a model family. In contrast, NAS-based architectures like \method \ automatically discover scaling strategies through grid-based search. We analyze two families of RF-DETR models derived from distinct scaling strategies: one based on a DINOv2-S backbone and another based on a DINOv2-B backbone. To evaluate how well each family scales, we compare their NAS-generated Pareto curves against those of D-FINE. Specifically, at each D-FINE size, we identify the RF-DETR variant with the same backbone that maximizes performance at a comparable latency. For example, when comparing to D-FINE (small), we select the RF-DETR model that offers the best accuracy without exceeding D-FINE (small)'s latency. Note that these \method models are different than those reported in Tables \ref{tab:coco_det} and \ref{tab:coco_det_scale}.

As shown in Table \ref{tab:scaling}, the DINOv2-S backbone family initially surpasses D-FINE in mAP@50:95 but fails to maintain this advantage at larger model sizes, suggesting that its scaling strategy is less effective than D-FINE's manual design. In contrast, the DINOv2-B backbone family shows the opposite trend, where the performance gap between D-FINE and \method \ narrows as latency increases. This implies that at higher latencies, the DINOv2-B based \method \ models could surpass D-FINE (and indeed RF-DETR (2x-large) outperforms D-FINE on mAP 50:95). Importantly, expanding the D-FINE model family would require substantial additional engineering effort, whereas extending the \method \ model family is straightforward; higher-latency variants can be sampled directly from the same NAS search without re-training. We present the COCO and RF100-VL results of our larger variants in Tables \ref{tab:coco_det_scale}, \ref{tab:coco_seg_scale}, and \ref{tab:rf100vl-scale}. We also include an \method \ Max variant on each dataset to show our method's maximum performance with latency less than 100ms, a scale other model families don't reach.

{
\setlength{\tabcolsep}{3.7em}
\begin{table}[t]
\caption{\textbf{mAP@50:95 Gap of \method \ vs D-FINE at Similar Latencies} We compare how different \method \ model families scale relative to D-FINE. D-FINE (nano) is excluded since it was not pretrained on Objects-365 and is therefore not expected to follow similar scaling trends. For each \method \ backbone, we select the highest accuracy Pareto-optimal NAS-mined model with latency up to that of the corresponding D-FINE variant. Notably, \method \ (DINOv2-B) achieves better scalability than \method \ (DINOv2-S) and D-FINE. Note that none of the RF-DETR models for COCO are finetuned.}
\label{tab:scaling}
\scalebox{0.7}{
\begin{tabular}{|l|c|c|c|c|}\hline
\rowcolor{gray!10}
\hline
\textbf{Method (Backbone)}& \textbf{S} & \textbf{M} & \textbf{L} & \textbf{XL} \\
\hline
D-FINE  \citep{peng2024dfine}      & 50.6 & 55.4 & 57.2 & 59.3  \\ \hline
\method \ (DINOv2-S) & +2.3 & +0.9 & -0.4 & -1.1 \\ \hline 
\method \ (DINOv2-B) & -3.1 & -1.3 & -1.2 & -0.7 \\ \hline
\end{tabular}
}
\end{table}
}

{
\setlength{\tabcolsep}{0.65em}
\begin{table}[t]
\caption{\textbf{COCO Detection Evaluation for Larger Model Variants.} We present \method's performance for L, XL, and 2XL sizes on COCO below. Notably, D-FINE (x-large) outperforms \method \ (x-large) on mAP 50:95. However, \method \ (2x-large) beats D-FINE by 0.8 AP, and is the first real-time detector to surpass 60 AP on COCO.}
\label{tab:coco_det_scale}
\scalebox{0.7}{
\begin{tabular}{|lcccccccccc|}
\hline
\rowcolor{gray!10}
\textbf{Model} & \multicolumn{1}{c|}{\textbf{Size}} & \textbf{\# Params.} & \textbf{GFLOPS} & \multicolumn{1}{c|}{\textbf{Latency (ms)}} & \textbf{AP} & \textbf{AP$_{50}$} & \textbf{AP$_{75}$} & \textbf{AP$_S$} & \textbf{AP$_M$} & \textbf{AP$_L$} \\ \hline
\rowcolor{gray!10}
\multicolumn{11}{|l|}{\textbf{Real-Time Object Detection w/ NMS}}  \\ \hline
\multicolumn{1}{|l|}{YOLOv8 \citep{yolov8}}     & \multicolumn{1}{c|}{L}     &   43.7M    &    165.2    & \multicolumn{1}{c|}{8.0}             &   49.5   &   64.7   &  54.0    &   30.2   &   55.1   &   68.5   \\ \hline
\multicolumn{1}{|l|}{YOLOv11 \citep{yolov11}}     & \multicolumn{1}{c|}{L}     &   25.3M    &    86.9    & \multicolumn{1}{c|}{6.5}             &  49.9   &   64.9   &   54.5   &  30.4    &   55.9   &   68.1   \\ \hline \hline
\multicolumn{1}{|l|}{YOLOv8 \citep{yolov8}}     & \multicolumn{1}{c|}{XL}     &   68.2M    &     257.8   & \multicolumn{1}{c|}{11.3}             &  50.5   &   65.6   &   55.1   &   30.0   &   56.2   &   69.5   \\ \hline
\multicolumn{1}{|l|}{YOLOv11 \citep{yolov11}}     & \multicolumn{1}{c|}{XL}     &   56.9M    &    194.9    & \multicolumn{1}{c|}{10.5}             &   50.9  &  66.1    &   55.4   &   31.5   &   56.6   &   68.7   \\ \hline \hline
\rowcolor{gray!10}
\multicolumn{11}{|l|}{\textbf{End-to-End Real-Time Object Detection}}                                                                \\ \hline
\multicolumn{1}{|l|}{RT-DETR \citep{zhao2024rtdetr}}      & \multicolumn{1}{c|}{R50}     &   42M    &    136    & \multicolumn{1}{c|}{8.5}             &      55.0   &   73.3   &  59.8 & 37.9  &  59.7    &   71.6   \\ \hline
\multicolumn{1}{|l|}{LW-DETR \citep{chen2024lw}}      & \multicolumn{1}{c|}{L}     &   46.8M    &     137.5     & \multicolumn{1}{c|}{6.9}             &   56.1  &   74.6   &   61.0   &   37.1   &  60.4    &   73.0   \\ \hline
\multicolumn{1}{|l|}{D-FINE \citep{peng2024dfine}}      & \multicolumn{1}{c|}{L}     &    31M   &    91    & \multicolumn{1}{c|}{7.5}             &  57.2   &   74.9   &   62.2   &   40.6   &    61.4  &   73.7   \\ \hline
\multicolumn{1}{|l|}{\method \ (Ours)}     & \multicolumn{1}{c|}{L}     &   33.9M    &    125.6    & \multicolumn{1}{c|}{6.8}             &   56.5  &  75.1    &   61.3   &   39.0   &  61.0    &   73.9   \\ \hline \hline
\multicolumn{1}{|l|}{RT-DETR \citep{zhao2024rtdetr}}      & \multicolumn{1}{c|}{R101}     &   76M    &    259    & \multicolumn{1}{c|}{12.0}             &   56.1  &   74.5   &   61.1   &   38.1   &   60.4   &   73.4   \\ \hline
\multicolumn{1}{|l|}{LW-DETR \citep{chen2024lw}}      & \multicolumn{1}{c|}{XL}     &    118.0M   &   342.5      & \multicolumn{1}{c|}{13.0}             &  58.3   &  76.9    &   63.3   &   40.2   &  63.3    &   74.7   \\ \hline
\multicolumn{1}{|l|}{D-FINE \citep{peng2024dfine}}      & \multicolumn{1}{c|}{XL}     &   62M    &    202    & \multicolumn{1}{c|}{11.5}             &  59.3   &   76.8   &   64.6   &  42.1    &  64.2    &   76.3   \\ \hline
\multicolumn{1}{|l|}{\method \ (Ours)}     & \multicolumn{1}{c|}{XL}     &   126.4M    &    299.3    & \multicolumn{1}{c|}{11.5}             &   58.6  &    77.4  &   63.8   &   40.3   &  63.9    &   76.2   \\ \hline \hline
\multicolumn{1}{|l|}{\method \ (Ours)}     & \multicolumn{1}{c|}{2XL}     &    126.9M   &    438.4    & \multicolumn{1}{c|}{17.2}             &  60.1   &   78.5   &   65.5   &   43.2   &   64.9   &   76.2   \\ \hline \hline
\multicolumn{1}{|l|}{\method \ (Ours)}     & \multicolumn{1}{c|}{Max}     &   132.4M    &    1742.5    & \multicolumn{1}{c|}{98.0}             &  61.8   &   79.7   &   67.7   &  47.5    &   66.1   &   76.0   \\ \hline
\end{tabular}
}
\end{table}
}

{
\setlength{\tabcolsep}{0.65em}
\begin{table}[t]
\caption{\textbf{COCO Segmentation Evaluation for Larger Model Variants.} We present \method's performance for L, XL, and 2XL sizes on the COCO segmentation benchmark below. We find that scaling up \method \ yields considerable performance improvements. In contrast, YOLOv8 and YOLOv11 do not significantly improve with scale.}
\label{tab:coco_seg_scale}
\scalebox{0.7}{
\begin{tabular}{|lcccccccccc|}
\hline
\rowcolor{gray!10}
\textbf{Model} & \multicolumn{1}{c|}{\textbf{Size}} & \textbf{\# Params.} & \textbf{GFLOPS} & \multicolumn{1}{c|}{\textbf{Latency (ms)}} & \textbf{AP} & \textbf{AP$_{50}$} & \textbf{AP$_{75}$} & \textbf{AP$_S$} & \textbf{AP$_M$} & \textbf{AP$_L$} \\ \hline
\rowcolor{gray!10}
\multicolumn{11}{|l|}{\textbf{Real-Time Instance Segmentation w/ NMS}}  \\ \hline
\multicolumn{1}{|l|}{YOLOv8 \citep{yolov8}}     & \multicolumn{1}{c|}{L}     &   46.0M    &    220.5    & \multicolumn{1}{c|}{9.7}             &  39.0   &   60.5   &   41.7   &   18.0   &   44.7   &   57.8   \\ \hline
\multicolumn{1}{|l|}{YOLOv11 \citep{yolov11}}     & \multicolumn{1}{c|}{L}     &   27.6M    &    132.2    & \multicolumn{1}{c|}{8.3}             &  39.5   &   61.5   &  42.1    &  18.6    &   45.5   &   59.4   \\ \hline \hline
\multicolumn{1}{|l|}{YOLOv8 \citep{yolov8}}     & \multicolumn{1}{c|}{XL}     &   71.8M    &    344.1    & \multicolumn{1}{c|}{14.0}             &  39.5   &   61.3   &   42.1   &   18.9   &   45.6   &   58.8   \\ \hline
\multicolumn{1}{|l|}{YOLOv11 \citep{yolov11}}     & \multicolumn{1}{c|}{XL}     &    62.1M   &     296.4   & \multicolumn{1}{c|}{13.7}             &   40.1  &   62.4   &  42.6    &   18.8   &   46.4   &   60.1   \\ \hline \hline
\rowcolor{gray!10}
\multicolumn{11}{|l|}{\textbf{End-to-End Real-Time Instance Segmentation}}                                                                \\ \hline
\multicolumn{1}{|l|}{\method \ (Ours)}     & \multicolumn{1}{c|}{L}     &    36.2M   &   151.1     & \multicolumn{1}{c|}{8.8}             &   47.1  &   70.5   &   50.9   &   28.4   &  52.1    &   65.6   \\ \hline
\multicolumn{1}{|l|}{\method \ (Ours)}     & \multicolumn{1}{c|}{XL}     &    38.1M   &    260.0    & \multicolumn{1}{c|}{13.5}             &  48.8   &   72.2   &   53.1   &   30.6   &   53.3   &   65.9    \\ \hline
\multicolumn{1}{|l|}{\method \ (Ours)}     & \multicolumn{1}{c|}{2XL}     &   38.6M    &    435.3    & \multicolumn{1}{c|}{21.8}             &   49.9  &   73.1   &   54.5   &   33.9   &   54.1   &   65.7   \\ \hline
\multicolumn{1}{|l|}{\method \ (Ours)}     & \multicolumn{1}{c|}{Max}     &    40.1M   &    1668.2    & \multicolumn{1}{c|}{95.6}             &  50.5   &   74.0   &   55.4   &   34.6   &   54.2   &   65.4   \\ \hline
\end{tabular}
}
\end{table}
}

{
\setlength{\tabcolsep}{0.65em}
\begin{table}[t]
\caption{\textbf{RF100-VL Detection Evaluation for Larger Model Variants.} We present \method's performance for L, XL, and 2XL sizes on RF100-VL below. Notably, \method \ (x-large) beats D-FINE by 0.5 AP. Fine-tuning \method \ (x-large) improves performance by an additional 0.4 AP.}
\label{tab:rf100vl-scale}
\scalebox{0.7}{
\begin{tabular}{|lcccccccccc|}
\hline
\rowcolor{gray!10}
\textbf{Model} & \multicolumn{1}{c|}{\textbf{Size}} & \textbf{\# Params.} & \textbf{GFLOPS} & \multicolumn{1}{c|}{\textbf{Latency (ms)}} & \textbf{AP} & \textbf{AP$_{50}$} & \textbf{AP$_{75}$} & \textbf{AP$_S$} & \textbf{AP$_M$} & \textbf{AP$_L$} \\ \hline
\rowcolor{gray!10}
\multicolumn{11}{|l|}{\textbf{Real-Time Object Detection w/ NMS}}  \\ \hline
\multicolumn{1}{|l|}{YOLOv8 \citep{yolov8}}     & \multicolumn{1}{c|}{L}     &   43.7M    &    165.2     & \multicolumn{1}{c|}{7.9}             &  56.5   &  82.1    &   61.1   &   7.1   &   46.0   &    48.9  \\ \hline
\multicolumn{1}{|l|}{YOLOv11 \citep{yolov11}}     & \multicolumn{1}{c|}{L}    &   25.3M    &    86.9      & \multicolumn{1}{c|}{6.4}             &  56.5   &  82.2    &  61.0    &  6.4    &   45.5   &   49.0   \\ \hline \hline
\multicolumn{1}{|l|}{YOLOv8 \citep{yolov8}}     & \multicolumn{1}{c|}{XL}   &   68.2M    &     257.8      & \multicolumn{1}{c|}{11.2}             &  56.5   &  82.3    &  61.0    &   6.6   &   45.7   &   47.9   \\ \hline
\multicolumn{1}{|l|}{YOLOv11 \citep{yolov11}}     & \multicolumn{1}{c|}{XL}   &   56.9M    &    194.9      & \multicolumn{1}{c|}{10.3}             &  56.2   &   81.7   &   60.8   &  6.1    &   45.9   &   48.1   \\ \hline \hline
\rowcolor{gray!10}
\multicolumn{11}{|l|}{\textbf{End-to-End Real-Time Object Detection}}                                                                \\ \hline
\multicolumn{1}{|l|}{RT-DETR \citep{zhao2024rtdetr}}      & \multicolumn{1}{c|}{R50}     &   42M    &    136    & \multicolumn{1}{c|}{8.4}             &  61.7   &   87.7   &  66.9    &   38.1   &   57.1   &  69.4    \\ \hline
\multicolumn{1}{|l|}{LW-DETR \citep{chen2024lw}}      & \multicolumn{1}{c|}{L}     &   46.8M    &     137.5   & \multicolumn{1}{c|}{6.8}             &  61.5   &   87.4   &   67.0   &   37.1   &  56.4    &   69.0   \\ \hline
\multicolumn{1}{|l|}{D-FINE \citep{peng2024dfine}}      & \multicolumn{1}{c|}{L}     &     31M   &    91   & \multicolumn{1}{c|}{7.5}             & 61.6    &  86.4    &  67.2    &  37.8    &  56.5    &   70.1   \\ \hline
\multicolumn{1}{|l|}{\method \ (Ours)}     & \multicolumn{1}{c|}{L}     &   34.1M    &    119.1    & \multicolumn{1}{c|}{6.2}             &  62.0   &   88.1   &   67.3   &  36.9    &   57.1   &   70.2   \\ \hline
\multicolumn{1}{|l|}{\method \ (Ours) w/ Fine-Tuning}     & \multicolumn{1}{c|}{L}     &   34.1M    &    119.1     & \multicolumn{1}{c|}{6.2}             &  62.3   &   88.2   &   67.4   &   37.1   &   57.2   &  70.3    \\ \hline \hline
\multicolumn{1}{|l|}{RT-DETR \citep{zhao2024rtdetr}}      & \multicolumn{1}{c|}{R101}     &   76M    &    259   & \multicolumn{1}{c|}{11.9}             &  61.0   &  87.4    &  66.2    &  36.6    &  56.3    &  68.2    \\ \hline
\multicolumn{1}{|l|}{LW-DETR \citep{chen2024lw}}      & \multicolumn{1}{c|}{XL}     &    118.0M   &   342.5     & \multicolumn{1}{c|}{13.0}             &  62.1   &  87.9    &  67.6    &  37.4    &  57.1    &   70.2   \\ \hline
\multicolumn{1}{|l|}{D-FINE \citep{peng2024dfine}}      & \multicolumn{1}{c|}{XL}     &    59.3   &   76.8     & \multicolumn{1}{c|}{11.4}             &  62.2   &  86.9    &   68.0   &  37.6    &   57.4   &   69.7   \\ \hline
\multicolumn{1}{|l|}{\method \ (Ours)}     & \multicolumn{1}{c|}{XL}     &   35.0M    &    199.0    & \multicolumn{1}{c|}{9.7}             &   62.6   &  88.5   &  67.9   &   39.0   &   57.8   &  70.4  \\ \hline
\multicolumn{1}{|l|}{\method \ (Ours) w/ Fine-Tuning}     & \multicolumn{1}{c|}{XL}     &  35.0M    &    199.0      & \multicolumn{1}{c|}{9.7}             &   63.0  &   88.7   &   68.2   &   38.8   &   58.2   &   70.6   \\ \hline \hline
\multicolumn{1}{|l|}{\method \ (Ours)}     & \multicolumn{1}{c|}{2XL}     &   123.5M    &    410.2   & \multicolumn{1}{c|}{15.6}             &  63.3   &   88.9   &    69.0  &  38.7    &   58.2   &   71.6   \\ \hline
\multicolumn{1}{|l|}{\method \ (Ours) w/ Fine-Tuning}     & \multicolumn{1}{c|}{2XL}     &   123.5M    &    410.2    & \multicolumn{1}{c|}{15.6}             &   63.5  &   89.0   &   69.2   &   38.9   &   58.3   &   71.7   \\ \hline
\end{tabular}
}
\end{table}
}

\section{Per-Knob Sensitivity Analysis}
\label{appendix:tunable-knobs}

\begin{figure}
    \centering
    \includegraphics[width=0.48\linewidth]{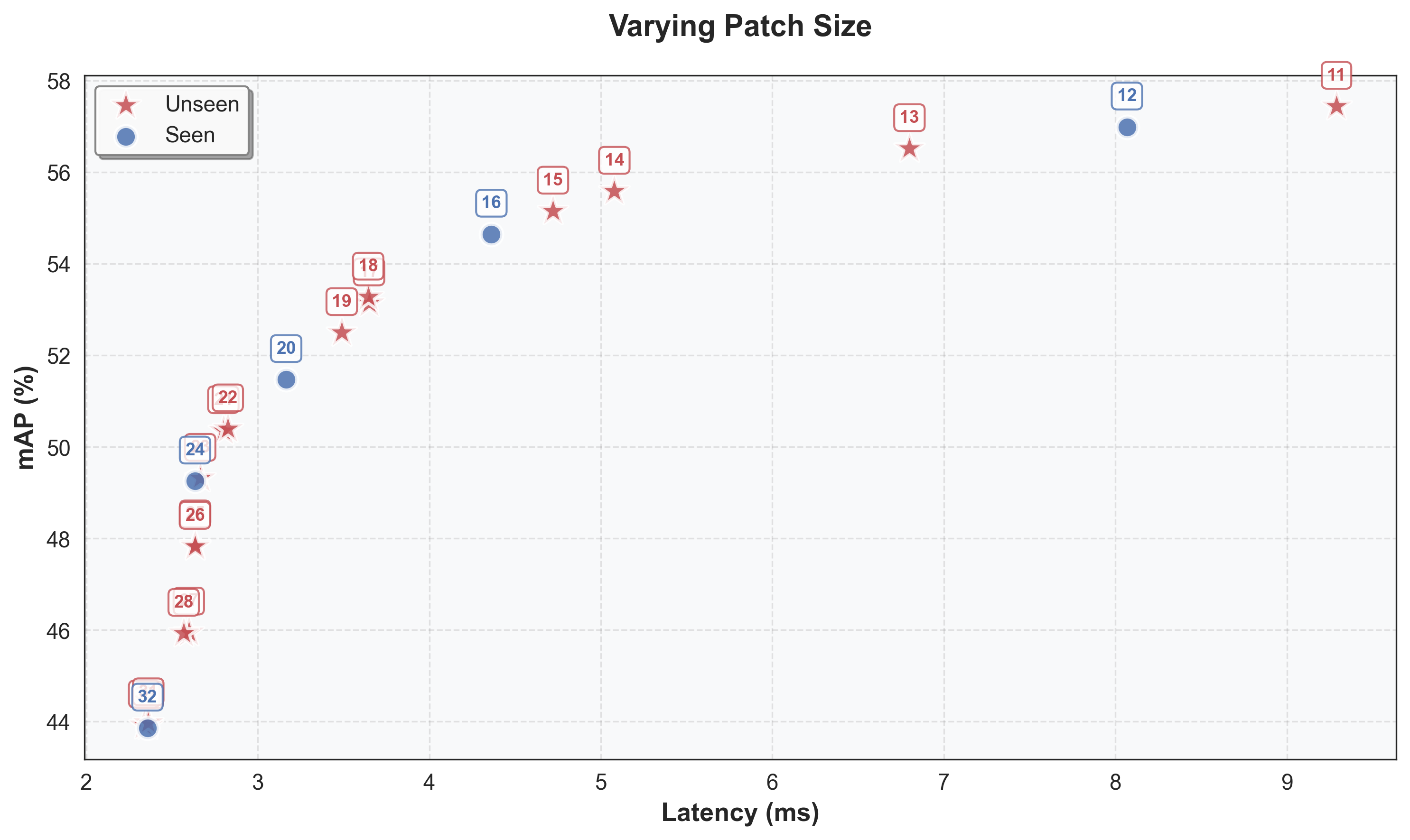}
    \includegraphics[width=0.48\linewidth]{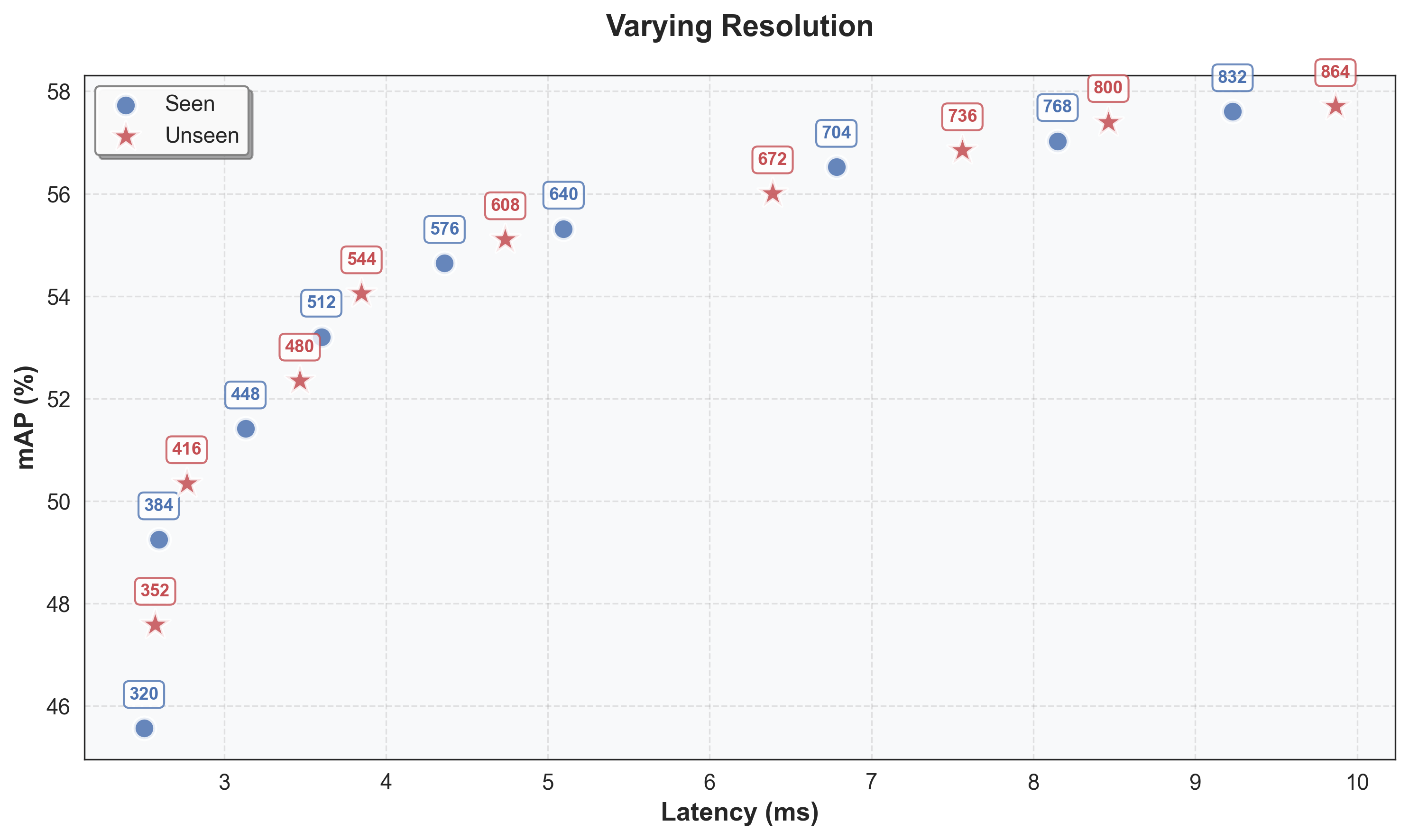}
    \caption{\textbf{Per Knob Sensitivity Analysis.} Despite never seeing certain resolutions and patch sizes, \method \ is able to gracefully interpolate to novel model configurations. }
    \label{fig:knob_sensitivity}
\end{figure}

We evaluate the impact of varying resolution and patch size in Figure \ref{fig:knob_sensitivity}. Both curves follow a clear Pareto frontier, and are consistent with findings from prior work like FlexiViT \citep{beyer2023flexivit}. \method \ is able to gracefully interpolate between seen (blue circles) and unseen (red stars) configurations during inference. Importantly, unseen configurations closely track the trend established by the seen configurations, demonstrating that RF-DETR generalizes beyond the model configurations encountered during training. 

\section{Impact of NAS Fine-Tuning on COCO}
\label{appendix:fine-tuning}
We find that fine-tuning after NAS provides limited benefit for COCO. We posit that the NAS ``architecture augmentation'' acts as a strong regularizer, and additional training without this regularization leads to degraded performance. Specifically, when models are pre-trained with strong regularization, removing the regularization during fine-tuning leads to overfitting. As shown in Tables \ref{tab:coco-det-ft} and \ref{tab:coco-seg-ft}, this trend is consistent across both detection and segmentation tasks. Interestingly, models trained on RF100-VL benefit more from fine-tuning, likely because they require more than 100 epochs to converge. In such cases, we posit that reducing the total number of NAS configurations during training, or training for more than 100 epochs with weight-sharing NAS can improve performance. 

{
\setlength{\tabcolsep}{0.65em}
\begin{table}[h]
\caption{\textbf{COCO Detection Fine-Tuning Evaluation.} We find that fine-tuning after NAS provides limited benefit for COCO detection, particularly for larger model sizes.} 
\label{tab:coco-det-ft}
\scalebox{0.7}{
\begin{tabular}{|lcccccccccc|}
\hline
\rowcolor{gray!10}
\textbf{Model} & \multicolumn{1}{c|}{\textbf{Size}} & \textbf{\# Params.} & \textbf{GFLOPS} & \multicolumn{1}{l|}{\textbf{Latency (ms)}} & \textbf{AP} & \textbf{AP$_{50}$} & \textbf{AP$_{75}$} & \textbf{AP$_S$} & \textbf{AP$_M$} & \textbf{AP$_L$} \\ \hline
\rowcolor{gray!10}
\multicolumn{11}{|l|}{\textbf{End-to-End Real-Time Object Detectors}}                                                                \\ \hline
\multicolumn{1}{|l|}{\method \ (Ours)}     & \multicolumn{1}{c|}{N}     &   30.5M    &   31.9    & \multicolumn{1}{c|}{2.3}             &  48.0  &   67.0   &   51.4   &  25.2   &   53.5  &   70.0  \\ \hline
\multicolumn{1}{|l|}{\method \ (Ours) w/ Fine-Tuning}     & \multicolumn{1}{c|}{N}     &       30.5M    &   31.9        & \multicolumn{1}{c|}{2.3}             &  +0.4  &   +0.6   &   +0.3   &  +0.1   &  +0.1  &   +1.3  \\ \hline \hline
\multicolumn{1}{|l|}{\method \ (Ours)}     & \multicolumn{1}{c|}{S}     &    32.1M   &   59.8     & \multicolumn{1}{c|}{3.5}             &  52.9  &  71.9    &   57.0   &  32.0   &  58.3   &  73.0   \\ \hline 
\multicolumn{1}{|l|}{\method \ (Ours) w/ Fine-Tuning}     & \multicolumn{1}{c|}{S}     &       32.1M   &   59.8        & \multicolumn{1}{c|}{3.5}             &  +0.1  &  +0.2    &   +0.2   &  -0.2   &  +0.2   &  +0.1   \\ \hline \hline
\multicolumn{1}{|l|}{\method \ (Ours)}     & \multicolumn{1}{c|}{M}     &  33.7M     &   78.8     & \multicolumn{1}{c|}{4.4}             &  54.7  &   73.5   &   59.2   &  36.1   &  59.7   &  73.8   \\ \hline 
\multicolumn{1}{|l|}{\method \ (Ours) w/ Fine-Tuning}     & \multicolumn{1}{c|}{M}     &       33.7M     &   78.8        & \multicolumn{1}{c|}{4.4}             &  +0.0  &   +0.1   &   +0.0   &  -0.1   &  +0.1   &  -0.1   \\ \hline \hline
\multicolumn{1}{|l|}{\method \ (Ours)}     & \multicolumn{1}{c|}{L}     &   33.9M    &    125.6    & \multicolumn{1}{c|}{6.8}             &   56.5  &  75.1    &   61.3   &   39.0   &  61.0    &   73.9   \\ \hline
\multicolumn{1}{|l|}{\method \ (Ours) w/ Fine-Tuning}     & \multicolumn{1}{c|}{L}     &    33.9M   &   125.6     & \multicolumn{1}{c|}{6.8}             &   +0.0  &   +0.0   &   +0.0   &   -0.1   &   +0.1  &  +0.1    \\ \hline \hline
\multicolumn{1}{|l|}{\method \ (Ours)}     & \multicolumn{1}{c|}{XL}     &   126.4M    &    299.3    & \multicolumn{1}{c|}{11.5}             &   58.6  &    77.4  &   63.8   &   40.3   &  63.9    &   76.2   \\ \hline
\multicolumn{1}{|l|}{\method \ (Ours) w/ Fine-Tuning}     & \multicolumn{1}{c|}{XL}     &    126.4M   &   299.3     & \multicolumn{1}{c|}{11.5}             &  +0.3   &  +0.1    &   +0.2   &   +0.5   &   +0.4   &   +0.1   \\ \hline \hline
\multicolumn{1}{|l|}{\method \ (Ours)}     & \multicolumn{1}{c|}{2XL}     &    126.9M   &    438.4    & \multicolumn{1}{c|}{17.2}             &  60.1   &   78.5   &   65.5   &   43.2   &   64.9   &   76.2   \\ \hline
\multicolumn{1}{|l|}{\method \ (Ours) w/ Fine-Tuning}     & \multicolumn{1}{c|}{2XL}     &    126.9M   &   438.4     & \multicolumn{1}{c|}{17.2}             &  +0.1   &   +0.0   &   +0.3   &   +0.5   &   +0.2   &   +0.1   \\ \hline
\end{tabular}
}
\end{table}
}

{
\setlength{\tabcolsep}{0.67em}
\begin{table}[h]
\caption{\textbf{COCO Segmentation Fine-Tuning Evaluation.} We find that fine-tuning after NAS provides limited benefit for COCO segmentation, particularly for larger model sizes.} 
\label{tab:coco-seg-ft}
\scalebox{0.7}{
\begin{tabular}{|lcccccccccc|}
\hline
\rowcolor{gray!10}
\textbf{Model} & \multicolumn{1}{c|}{\textbf{Size}} & \textbf{\# Params.} & \textbf{GFLOPS} & \multicolumn{1}{l|}{\textbf{Latency (ms)}} & \textbf{AP} & \textbf{AP$_{50}$} & \textbf{AP$_{75}$} & \textbf{AP$_S$} & \textbf{AP$_M$} & \textbf{AP$_L$} \\ \hline
\rowcolor{gray!10}
\multicolumn{11}{|l|}{\textbf{End-to-End Real-Time Object Detectors}}                                                                \\ \hline
\multicolumn{1}{|l|}{\method-Seg. (Ours)}     & \multicolumn{1}{c|}{N}     &   33.6M    &    50.0    & \multicolumn{1}{c|}{3.4}             &  40.3  &  63.0    &  42.6    &  16.3   &  45.3   &   63.6  \\ \hline 
\multicolumn{1}{|l|}{\method-Seg.  w/ Fine-Tuning (Ours)}     & \multicolumn{1}{c|}{N}     &   33.6M    &   50.0     & \multicolumn{1}{c|}{3.4}             &  +0.1  &  +0.4   &  +0.0   &  -0.5 &  +0.2 &  +0.7  \\ \hline \hline
\multicolumn{1}{|l|}{\method-Seg. (Ours)}     & \multicolumn{1}{c|}{S}     &   33.7M   &    70.6    & \multicolumn{1}{c|}{4.4}             &  43.1  &   66.2   & 45.9     &  21.9   &  48.5   &   64.1  \\ \hline
\multicolumn{1}{|l|}{\method \ w/ Fine-Tuning (Ours)}     & \multicolumn{1}{c|}{S}     &  Did     &  Not     & \multicolumn{1}{c|}{Improve}             &  -    &   -   &   -   &   -   &   -   &   -   \\ \hline \hline
\multicolumn{1}{|l|}{\method-Seg. (Ours)}     & \multicolumn{1}{c|}{M}     &    35.7M      &    102.0    & \multicolumn{1}{c|}{5.9}             &  45.3  &   68.4   &  48.8    &   25.5  &   50.4  &  65.3   \\ \hline
\multicolumn{1}{|l|}{\method \ w/ Fine-Tuning (Ours)}     & \multicolumn{1}{c|}{M}     & Did     &  Not     & \multicolumn{1}{c|}{Improve}             &  -    &   -   &    -  &   -   &   -   &   -   \\ \hline \hline
\multicolumn{1}{|l|}{\method \ (Ours)}     & \multicolumn{1}{c|}{L}     &    36.2M   &   151.1     & \multicolumn{1}{c|}{8.8}             &   47.1  &   70.5   &   50.9   &   28.4   &  52.1    &   65.6   \\ \hline
\multicolumn{1}{|l|}{\method \ (Ours) w/ Fine-Tuning}     & \multicolumn{1}{c|}{L}     & Did     &  Not     & \multicolumn{1}{c|}{Improve}            &   -   &   -   &    -  &   -   &  -    &   -   \\ \hline \hline
\multicolumn{1}{|l|}{\method \ (Ours)}     & \multicolumn{1}{c|}{XL}     &    38.1M   &    260.0    & \multicolumn{1}{c|}{13.5}             &  48.8   &   72.2   &   53.1   &   30.6   &   53.3   &   65.9    \\ \hline
\multicolumn{1}{|l|}{\method \ (Ours) w/ Fine-Tuning}     & \multicolumn{1}{c|}{XL}     &  Did     &  Not     & \multicolumn{1}{c|}{Improve}            &   -   &  -    &   -   &   -   &  -    &   -   \\ \hline \hline
\multicolumn{1}{|l|}{\method \ (Ours)}     & \multicolumn{1}{c|}{2XL}     &   38.6M    &    435.3    & \multicolumn{1}{c|}{21.8}             &   49.9  &   73.1   &   54.5   &   33.9   &   54.1   &   65.7   \\ \hline
\multicolumn{1}{|l|}{\method \ (Ours) w/ Fine-Tuning}     & \multicolumn{1}{c|}{2XL}     &   Did     &  Not     & \multicolumn{1}{c|}{Improve}     &   -   &   -   &   -   &   -   &    -  &   -   \\ \hline
\end{tabular}
}
\end{table}
}
\section{Impact of Dataset Characteristics on Tunable Knobs}

\begin{figure}
    \centering
    \includegraphics[width=0.48\linewidth]{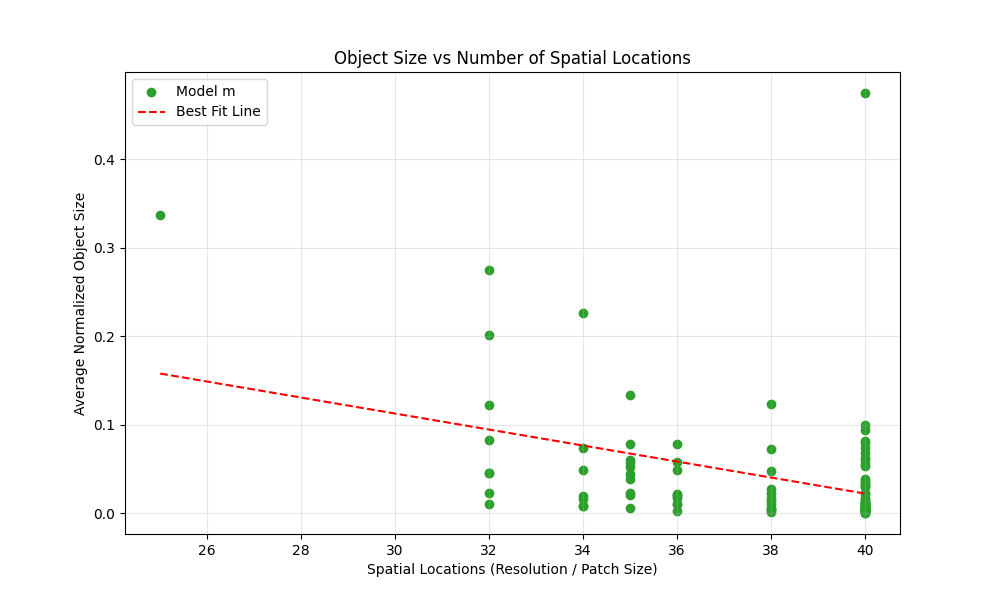}
    \includegraphics[width=0.48\linewidth]{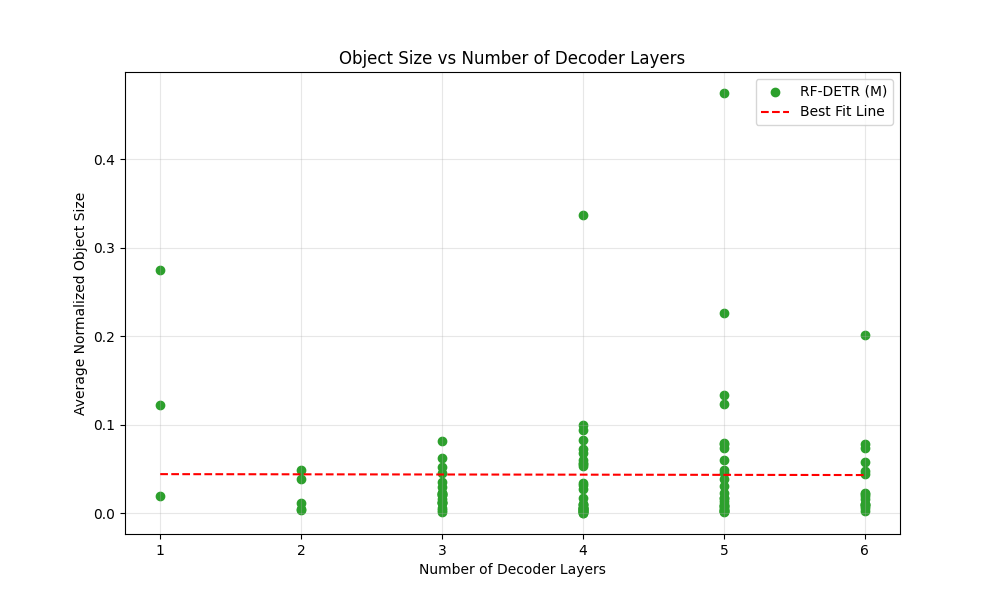}\\
    \includegraphics[width=0.48\linewidth]{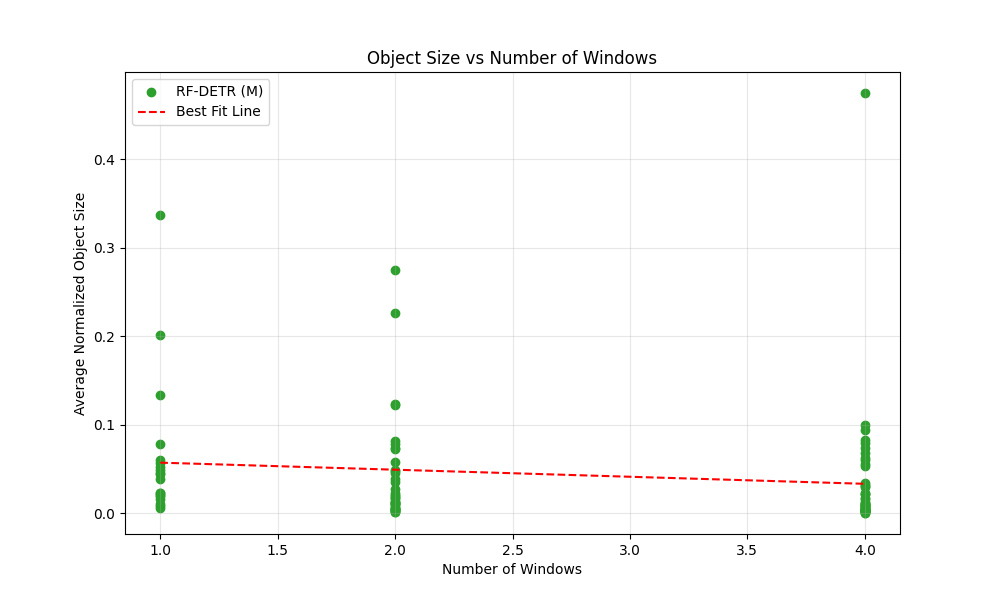}
    \includegraphics[width=0.48\linewidth]{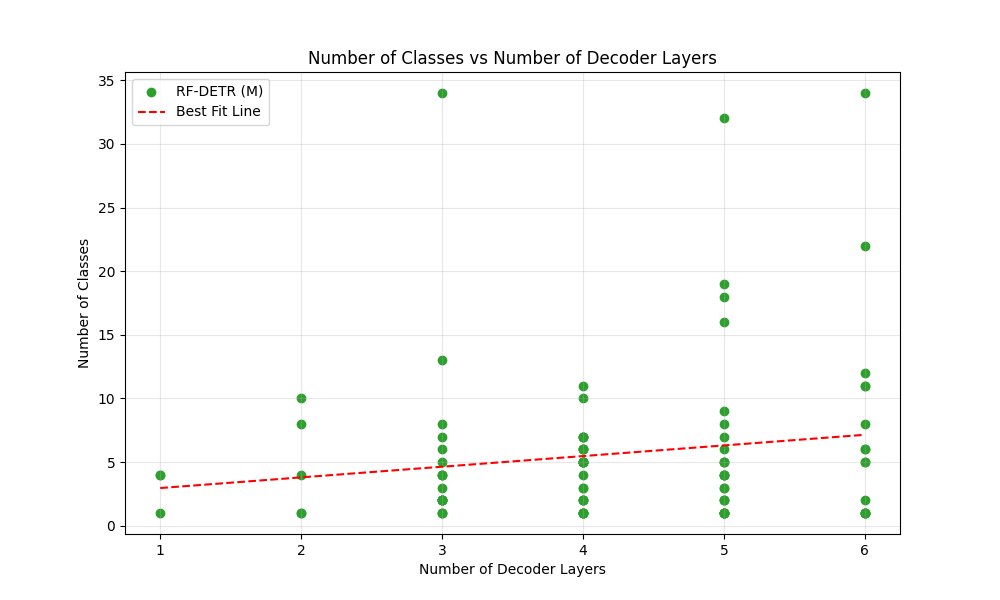} \\
     \includegraphics[width=0.48\linewidth]{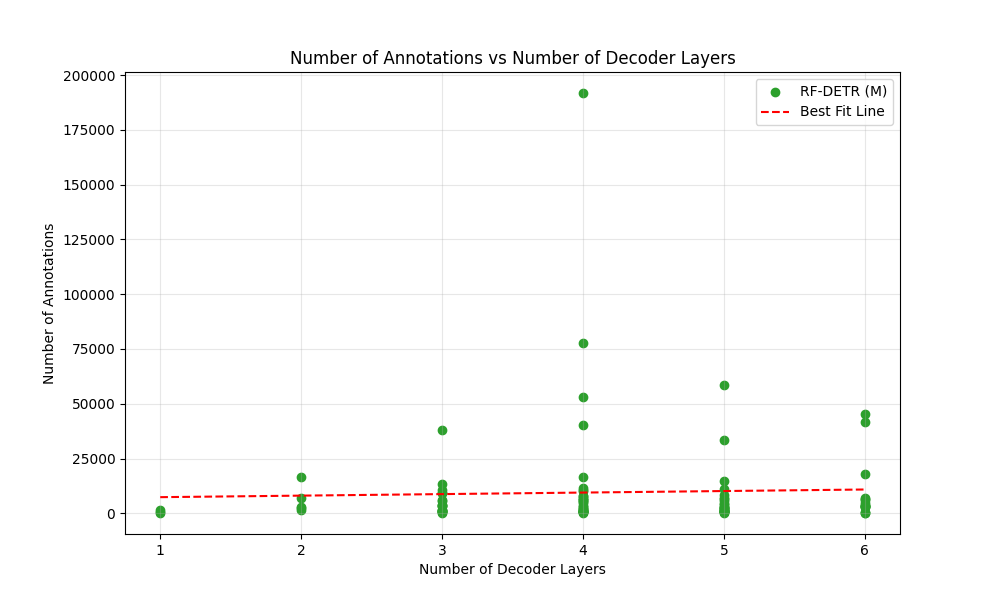} 
    \includegraphics[width=0.48\linewidth]{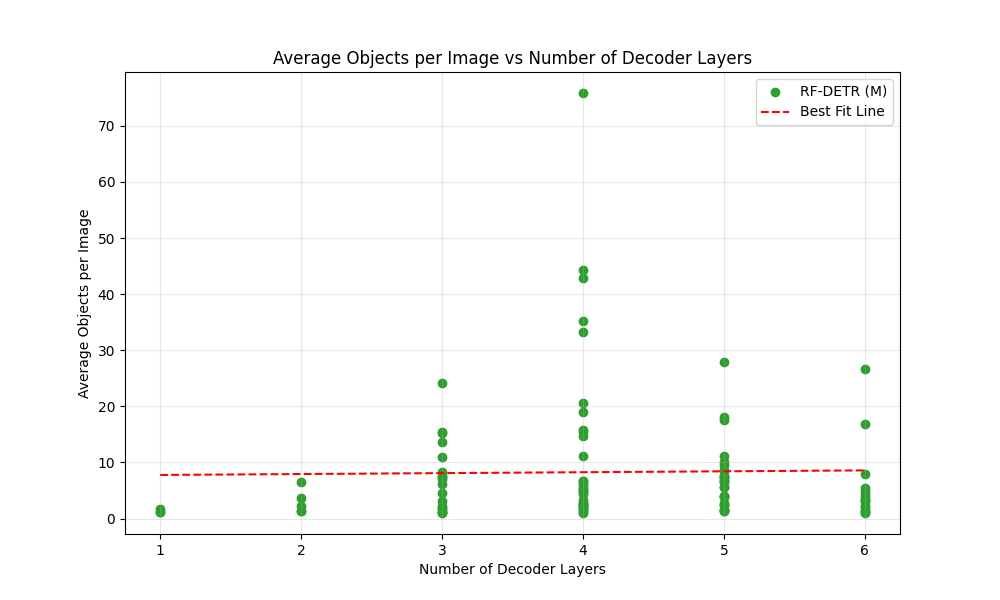} \\
    \includegraphics[width=0.48\linewidth]{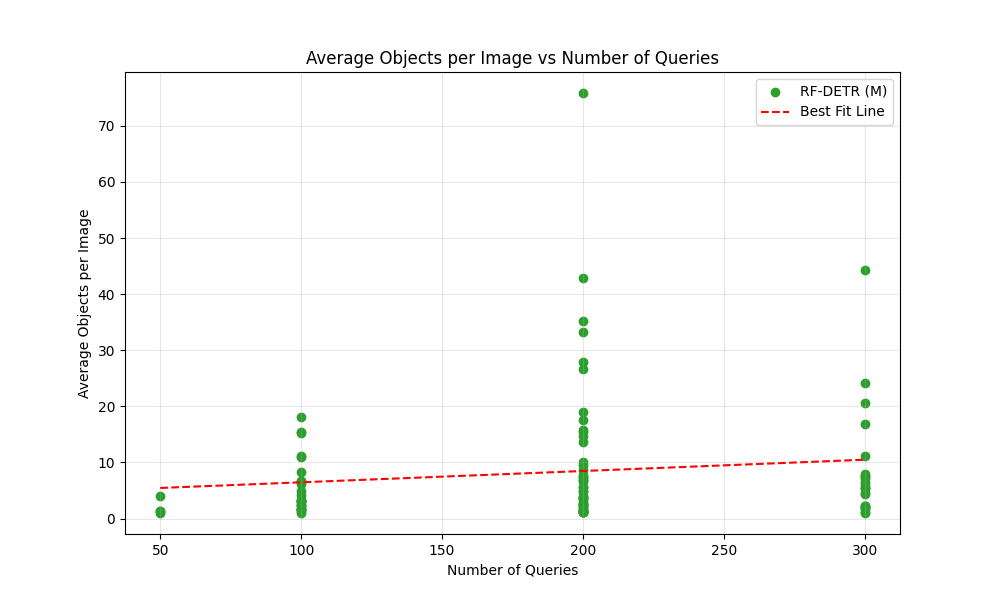}
    \includegraphics[width=0.48\linewidth]{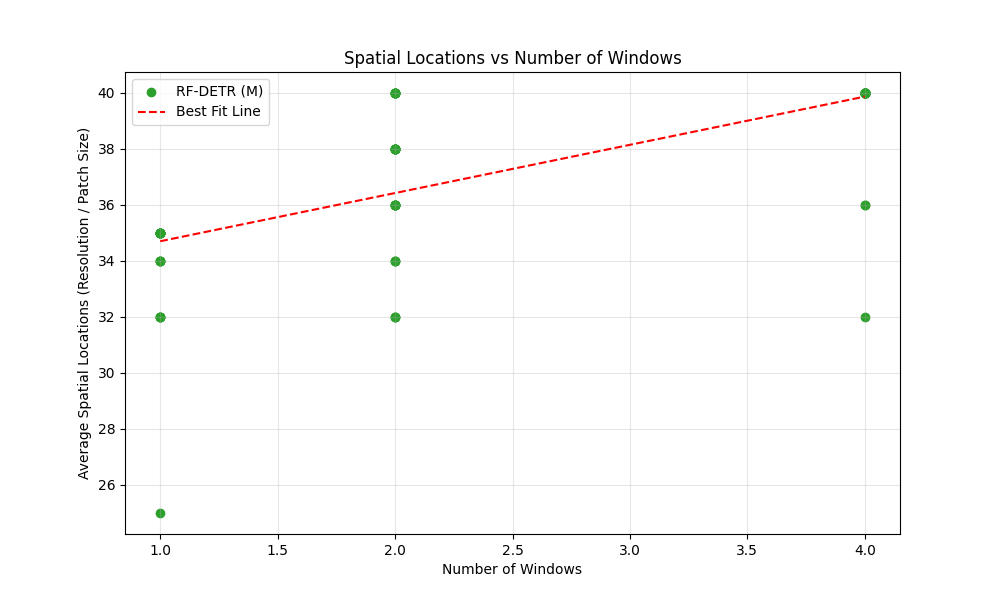}
    \caption{\textbf{Impact of Dataset Characteristics on Tunable Knobs}. We visualize the correlation between key dataset characteristics and several tunable architectural knobs above. Across all subplots, individual points represent different datasets within RF100-VL while dashed lines show corresponding linear trends. Overall, the results indicate that object-centric properties of datasets such as average object size, number of classes, object density, and total annotations tend to have only modest influence on architectural choices like the number of decoder layers, number of windows, number of queries, and spatial resolution. Slight positive or negative trends appear in some cases (e.g., more classes or more objects per image loosely correlating with deeper decoders or higher query counts), but the scatter remains wide, suggesting no strong deterministic relationship. These findings highlight that while dataset characteristics offer some intuition for selecting model hyperparameters, optimal configurations ultimately depend on a combination of factors rather than any single dataset attribute.}
    \label{fig:dataset_char}
\end{figure}

We evaluate the impact of different dataset characteristics on optimal network configurations for \method \ (medium) on RF100-VL in Table \ref{tab:regression-results} and Figure \ref{fig:dataset_char}. We compare different combinations of object size, number of spatial locations, number of decoder layers, number of windows, number of classes, number of annotations, objects per image, and number of queries below. We do not expect these relationships to be linear, but expect that they will be monotonic (e.g. non-zero slope).  For example, we find strong correlations between the number of classes and number of decoder layers, objects per image and number of queries, spatial locations and number of windows. Notably, we do not find strong correlations between object size and number of decoder layers, number of annotations and number of decoder layers, and objects per image and number of decoder layers

{
\setlength{\tabcolsep}{0.25em}
\begin{table}[h]
\caption{\textbf{Regression Analysis.} We evaluate the linear relationships between various dataset characteristics and model parameters. Although these correlations are non-linear, the line-of-best-fit helps explain general trends. P-values indicate the significance of the correlation.} 
\label{tab:regression-results}
\scalebox{0.85}{
\begin{tabular}{|lccccc|}
\hline
\rowcolor{gray!10}
\textbf{Relationship} & \textbf{Slope} & \textbf{Intercept} & \textbf{R-squared} & \textbf{P-value} & \textbf{Std. Error} \\ \hline
\multicolumn{1}{|l|}{Average Object Size vs Number of Spatial Locations} & -0.009 & 0.384 & 0.148 & 0.000 & 0.002 \\ \hline
\multicolumn{1}{|l|}{Average Object Size vs Number of Decoder Layers} & -0.000 & 0.044 & 0.000 & 0.971 & 0.006 \\ \hline
\multicolumn{1}{|l|}{Average Object Size vs Number of Windows} & -0.008 & 0.065 & 0.019 & 0.170 & 0.006 \\ \hline
\multicolumn{1}{|l|}{Number of Classes vs Number of Decoder Layers} & 0.837 & 2.125 & 0.026 & 0.106 & 0.513 \\ \hline
\multicolumn{1}{|l|}{Number of Annotations vs Number of Decoder Layers} & 698.763 & 6654.795 & 0.001 & 0.706 & 1848.733 \\ \hline
\multicolumn{1}{|l|}{Average Objects per Image vs Number Decoder Layers} & 0.163 & 7.618 & 0.000 & 0.857 & 0.904 \\ \hline
\multicolumn{1}{|l|}{Average Objects per Image vs Number of Queries} & 0.020 & 4.457 & 0.019 & 0.171 & 0.015 \\ \hline
\multicolumn{1}{|l|}{Number of Spatial Locations vs Number of Windows} & 1.722 & 32.984 & 0.492 & 0.000 & 0.177 \\ \hline
\end{tabular}
}
\end{table}
}

\section{Impact of Fixed Architecture on RF100-VL}
\begin{figure}[h]
\centering
\begin{minipage}[c]{0.5\linewidth}
    \includegraphics[width=\linewidth]{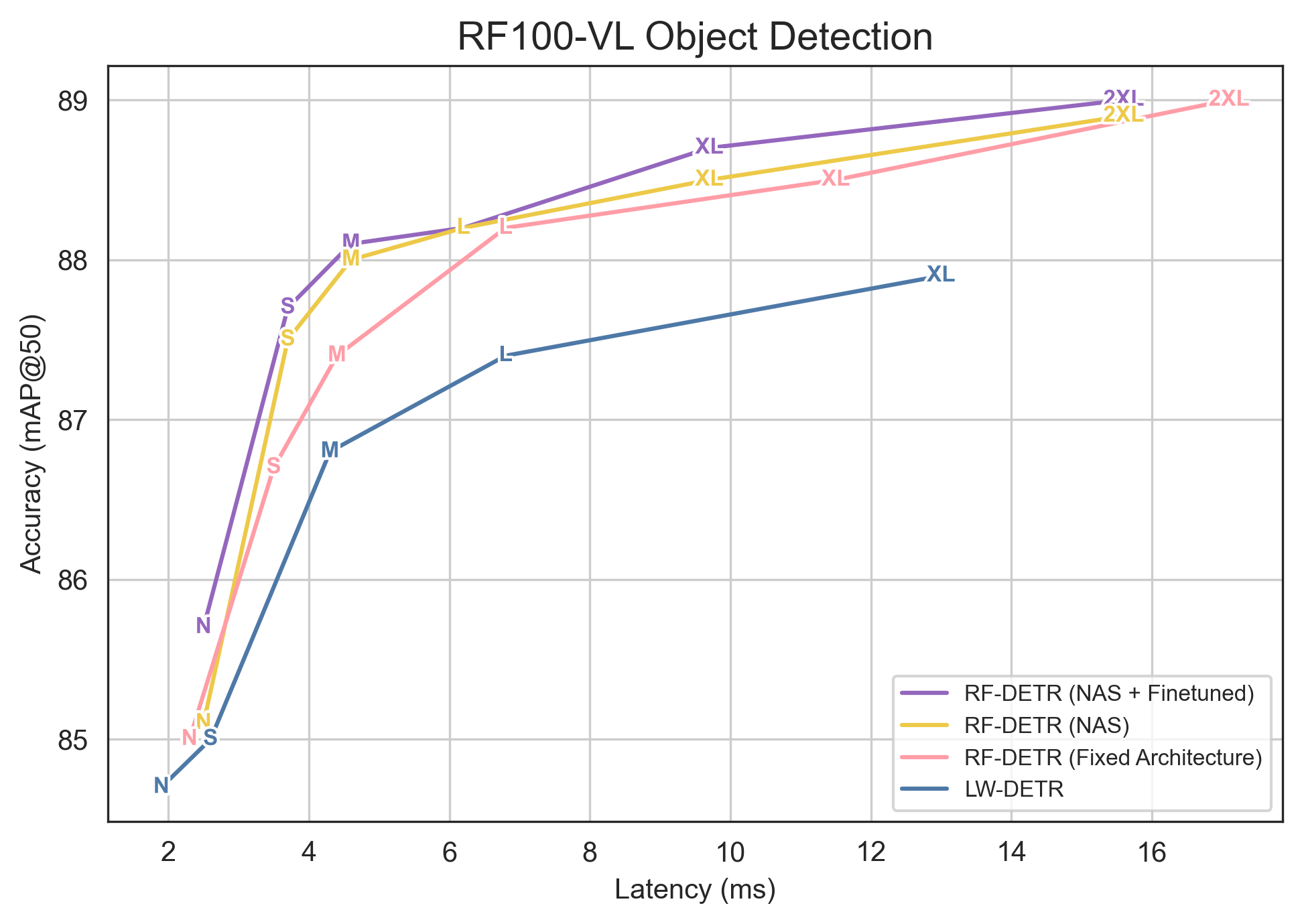}
\end{minipage}\hfill
\begin{minipage}[c]{0.48\linewidth}
    \caption{{\bf Ablating Fixed Architecture RF100-VL}. We evaluate the benefit of dataset-specific NAS by transferring the COCO-optimized \method \ architecture to RF100-VL. Although the fixed architecture was not tuned for RF100-VL, it still outperforms LW-DETR. Running NAS directly on RF100-VL further improves performance over the fixed architecture. Additional fine-tuning provides consistent gains across all model sizes, with particularly strong improvements for smaller models. This is consistent with our observations on COCO object detection.}
    \label{fig:fixed_arch_ablation}
\end{minipage}
\end{figure}

We evaluate the impact of transferring a NAS architecture optimized for COCO to RF100-VL in Table \ref{tab:rf100-vl-fixed-arch} and Figure \ref{fig:fixed_arch_ablation}. We find that these fixed architecture models perform remarkably well without further dataset-specific NAS. Specifically, \method \ (large) model with a fixed architecture achieves the best performance among all prior real-time models on COCO. However, dataset-specific NAS yields  significant improvements. Notably, the performance delta from LW-DETR to the fixed architecture is comparable to the improvement from the fixed architecture to the NAS-optimized model on the target dataset for nano, small, and medium scale models.

{
\setlength{\tabcolsep}{0.25em}
\begin{table}[h]
\caption{\textbf{RF100-VL Fixed Architecture Evaluation.} We evaluate the transfer of architectures optimized for COCO to RF100-VL. Fixed architecture models perform well without additional dataset-specific NAS, with the \method \ (large) model achieving the best performance among prior real-time models. However, dataset-specific NAS provides significant further gains.} 
\label{tab:rf100-vl-fixed-arch}
\scalebox{0.7}{
\begin{tabular}{|lcccccccccc|}
\hline
\rowcolor{gray!10}
\textbf{Model} & \multicolumn{1}{c|}{\textbf{Size}} & \textbf{\# Params.} & \textbf{GFLOPS} & \multicolumn{1}{l|}{\textbf{Latency (ms)}} & \textbf{AP} & \textbf{AP$_{50}$} & \textbf{AP$_{75}$} & \textbf{AP$_S$} & \textbf{AP$_M$} & \textbf{AP$_L$} \\ \hline
\rowcolor{gray!10}
\multicolumn{11}{|l|}{\textbf{End-to-End Real-Time Object Detectors}}                                                                \\ \hline
\multicolumn{1}{|l|}{\method \ (Ours) Fixed Architecture}     & \multicolumn{1}{c|}{N}     &   30.5M    &   31.9      & \multicolumn{1}{c|}{2.3}             & 57.7   &   85.0   &  61.9    &   30.8  &   51.5  &   67.4  \\ \hline 
\multicolumn{1}{|l|}{\method \ (Ours)}     & \multicolumn{1}{c|}{N}     &   31.2M   &   34.5     & \multicolumn{1}{c|}{2.5}             & 57.8   &   85.1   &   62.5   &  30.1   &   52.2   &   67.2  \\ \hline
\multicolumn{1}{|l|}{\method \ w/ Fine-Tuning (Ours)}     & \multicolumn{1}{c|}{N}     &   31.2M   &   34.5     & \multicolumn{1}{c|}{2.5}             &  58.6  &  85.7   &   63.0  &  31.0  &  53.2   &  67.6  \\ \hline \hline
\multicolumn{1}{|l|}{\method \ (Ours) Fixed Architecture}     & \multicolumn{1}{c|}{S}     &   32.1M   &   59.8      & \multicolumn{1}{c|}{3.5}             & 60.2   &  86.7    &   65.0   &  34.2   &   54.4  &   68.9  \\ \hline 
\multicolumn{1}{|l|}{\method \ (Ours)}     & \multicolumn{1}{c|}{S}     &    33.5M   &    62.4    & \multicolumn{1}{c|}{3.7}             & 60.9   &   87.5   &   66.1   &  34.2   &   55.7  &  69.6   \\ \hline
\multicolumn{1}{|l|}{\method \ w/ Fine-Tuning (Ours)}     & \multicolumn{1}{c|}{S}     &    33.5M   &    62.4    & \multicolumn{1}{c|}{3.7}             &  61.2  &  87.7    &  66.1   &  34.9  &  55.6  & 69.5   \\ \hline \hline
\multicolumn{1}{|l|}{\method \ (Ours) Fixed Architecture}     & \multicolumn{1}{c|}{M}     &  33.7M     &   78.8    & \multicolumn{1}{c|}{4.4}             & 61.2   &   87.4   &  66.4    &   35.8  &   56.1  &   69.8   \\ \hline
\multicolumn{1}{|l|}{\method \ (Ours)}     & \multicolumn{1}{c|}{M}     &   33.6M    &   91.0     & \multicolumn{1}{c|}{4.6}             & 61.7   &    88.0  &   66.9   &  35.8   &   56.5  &   70.0  \\ \hline
\multicolumn{1}{|l|}{\method \ w/ Fine-Tuning (Ours)}     & \multicolumn{1}{c|}{M}     &   33.6M    &   91.0     & \multicolumn{1}{c|}{4.6}             & 62.0   &  88.1   &  67.1   &  36.2   &  56.4  &  70.2   \\ \hline \hline

\multicolumn{1}{|l|}{\method \ (Ours) w/ Fixed Architecture}     & \multicolumn{1}{c|}{L}     &   33.9M    &    125.6   & \multicolumn{1}{c|}{6.8}             & 62.2   &   88.2   &  67.8    &   37.7  &   57.0  &   70.5  \\ \hline
\multicolumn{1}{|l|}{\method \ (Ours)}     & \multicolumn{1}{c|}{L}     &   34.1M    &    119.1    & \multicolumn{1}{c|}{6.2}             &  62.0   &   88.1   &   67.3   &  36.9    &   57.1   &   70.2   \\ \hline
\multicolumn{1}{|l|}{\method \ (Ours) w/ Fine-Tuning}     & \multicolumn{1}{c|}{L}     &   34.1M    &    119.1     & \multicolumn{1}{c|}{6.2}             &  62.3   &   88.2   &   67.4   &   37.1   &   57.2   &  70.3    \\ \hline \hline
\multicolumn{1}{|l|}{\method \ (Ours, DINOv2-Base) w/ Fixed Architecture}     & \multicolumn{1}{c|}{XL}     &   126.4M    &    299.3       & \multicolumn{1}{c|}{11.5}             & 62.9   &  88.5    &   68.6   &  37.0   &   57.5  &   71.3  \\ \hline 
\multicolumn{1}{|l|}{\method \ (Ours)}     & \multicolumn{1}{c|}{XL}     &   35.0M    &    199.0    & \multicolumn{1}{c|}{9.7}             &   62.6   &  88.5   &  67.9   &   39.0   &   57.8   &  70.4  \\ \hline
\multicolumn{1}{|l|}{\method \ (Ours) w/ Fine-Tuning}     & \multicolumn{1}{c|}{XL}     &  35.0M    &    199.0      & \multicolumn{1}{c|}{9.7}             &   63.0  &   88.7   &   68.2   &   38.8   &   58.2   &   70.6   \\ \hline \hline
\multicolumn{1}{|l|}{\method \ (Ours, DINOv2-Base) w/ Fixed Architecture}     & \multicolumn{1}{c|}{2XL}     &   126.9M   &    438.4   & \multicolumn{1}{c|}{17.1}             & 63.2   &   89.0   &  69.3    &   38.4  &   58.4  &   71.5   \\ \hline
\multicolumn{1}{|l|}{\method \ (Ours, DINOv2-Base)}     & \multicolumn{1}{c|}{2XL}     &     123.5M    &    410.2    & \multicolumn{1}{c|}{15.6}             &  63.3   &   88.9   &    69.0  &  38.7    &   58.2   &   71.6   \\ \hline
\multicolumn{1}{|l|}{\method \ (Ours, DINOv2-Base) w/ Fine-Tuning}     & \multicolumn{1}{c|}{2XL}     &     123.5M    &    410.2     & \multicolumn{1}{c|}{15.6}             &   63.5  &   89.0   &   69.2   &   38.9   &   58.3   &   71.7   \\ \hline
\end{tabular}
}
\end{table}
}
\section{Ablation on Backbone Architecture with RF20-VL}
In Table \ref{tab:backbone-rf20vl}, we reproduce our ablation on the impact of backbone architecture on downstream model performance (Table \ref{tab:backbone}) on RF20-VL. All trends from the main paper hold. 
{
\setlength{\tabcolsep}{0.6em}
\begin{table}[h]
\caption{\textbf{Ablation on Backbone RF20-VL.}} 
\label{tab:backbone-rf20vl}
\scalebox{0.7}{
\begin{tabular}{|lccccccccc|}
\hline
\rowcolor{gray!10}
\multicolumn{1}{|l|}{\textbf{LW-DETR (M) + Gentler Hyperparameters}} & \textbf{\# Params.} & \textbf{GFLOPS} & \multicolumn{1}{l|}{\textbf{Latency (ms)}} & \textbf{AP} & \textbf{AP$_{50}$} & \textbf{AP$_{75}$} & \textbf{AP$_S$} & \textbf{AP$_M$} & \textbf{AP$_L$} \\ \hline
\multicolumn{1}{|l|}{\quad w/ CAEv2 ViT/S-16-Truncated Backbone}     &  28.3M   & 83.7 &  \multicolumn{1}{c|}{4.3}             &  64.4  &  92.2    &   71.2   &  33.8   &  56.8   &  71.9   \\ \hline
\multicolumn{1}{|l|}{\quad w/ DINOv2 ViT/S-14 Backbone}     &      32.3M      &   78.2     &  \multicolumn{1}{c|}{4.6}             &  65.2  &  92.5    &   72.8   &  37.8   & 58.2    &  72.6  \\ \hline
\multicolumn{1}{|l|}{\quad w/ SigLIPv2 ViT/B-32 Backbone$^*$} & 105.1M     &     81.6   &  \multicolumn{1}{c|}{4.5}             & 62.2   & 91.0  & 67.8 & 28.9 & 54.1 & 70.3 \\ \hline
\multicolumn{1}{|l|}{\quad w/ SAM2 Hiera-S Backbone$^*$} & 44.0M &  109.1 &  \multicolumn{1}{c|}{11.2}   &  65.2  &  92.5  & 72.7  & 37.8  & 58.2  &  72.1  \\ \hline
\end{tabular}
}
\end{table}
}

\section{Analysis on Buffering}
\label{appendix:throtting}

We further analyze the impact of buffering on the relative ordering of inference speed in Table \ref{tab:buffering}. Notably, we find that buffering beyond 200ms does not change latency measurements. However, we acknowledge that adding a 200ms buffer after every forward pass considerably increases overall inference time. Future work should consider alternatives to buffering to address power throttling.

{
\setlength{\tabcolsep}{2.6em}
\begin{table}[h]
\centering
\caption{\textbf{Analysis on Buffering.} We evaluate models with different amounts of buffering between consecutive forward passes. We find that buffering beyond 200ms does not provide any additional stability to latency measurements.}
\label{tab:buffering}
\scalebox{0.7}{
\begin{tabular}{|l|c|c|c|c|c|}
\hline
\rowcolor{gray!10}
\textbf{Model} & \textbf{mAP} & \textbf{0 ms} & \textbf{200 ms} & \textbf{400 ms} & \textbf{800 ms} \\
\hline
YOLOv8 (M)     & 47.3 & 5.5 ms & 5.4 ms & 5.4 ms & 5.4 ms\\ \hline
YOLOv11 (M)    & 48.4 & 5.0 ms & 5.0 ms & 5.1 ms & 5.1 ms\\ \hline
RT-DETR (R18)  & 49.0 & 4.4 ms & 4.4 ms & 4.4 ms & 4.4 ms\\ \hline
LW-DETR (M)    & 52.6 & 4.5 ms & 4.3 ms & 4.3 ms & 4.3 ms\\ \hline
D-FINE (M)     & 54.9 & 5.7 ms & 5.4 ms & 5.4 ms & 5.4 ms \\ \hline
RF-DETR (M)    & 54.7 & 4.7 ms & 4.4 ms & 4.4 ms & 4.4 ms\\
\hline
\end{tabular}
}
\end{table}
}

\section{Discussion on Notable Discovered Architectures}
\label{appendix:trends}

Several trends emerge from our weight-sharing NAS. First, we note that all tunable ``knobs'' are used when defining Pareto-optimal model families, validating our search space. This suggests that expanding the search space may further improve downstream performance.

Across Pareto-optimal models, patch size is consistent within model families. For example, the optimal patch size for RF-DETRs \ with a DINOv2-S backbone is 16, RF-DETRs \ with a DINOv2-B backbone is 20, and \method-Segs with a DINOv2-S backbone is 12. Pareto-optimal models also jointly scale encoder and decoder compute: patch size, number of windows, and resolution impact the encoder, while decoder depth, and number of queries affect the decoder. For \method-Seg, scaling resolution impacts the segmentation head. We find that using 2 windows in the encoder is typically optimal and resolution scales within a model family as we increase latency. On COCO, \method \ scales decoder depth while keeping the number of queries fixed, while \method-Seg simultaneously scales both axes. This likely reflects a minimum viable segmentation head depth; to offset its latency, \method-Seg reduces its total number of queries, yielding a thin, deep decoder, in contrast to \method’s wide, shallow decoder.


Next, we find that \method's performance is more correlated with the total number of spatial locations (e.g. resolution divided by patch size) rather than resolution or patch size alone. Scaling resolution with a fixed patch size yields similar results to scaling patch size with a fixed resolution, since vision transformers are agnostic to absolute input resolution after the patchify-and-project operation. To verify this, we constructed an alternative model family with fixed resolution ($640$) and varied patch sizes to preserve the total number of spatial locations. Specifically, we evaluate \method \ (nano) with a patch size of $27$, \method \ (small) with a patch size of $21$, and \method \ (medium) with a patch size of $18$. Surprisingly, all model results are nearly identical to the Pareto-optimal family. Notably, patch sizes of $27$ and $18$ were unseen during training, demonstrating \method's strong generalization to novel patch sizes \citep{beyer2023flexivit}. However, we find that this trend does not hold for \method-Seg because segmentation features are always upsampled to $\frac{1}{4}$ of the input image resolution. As a result, scaling \method-Seg's input resolution affects both the number of spatial locations and the segmentation feature resolution. Specifically, \method-Seg (nano, small, medium) uses input resolutions of 312, 384, and 432 with patch size 12, yielding segmentation feature resolutions of 78, 96, and 108 and 26, 32, and 36 spatial locations, respectively. In contrast, increasing patch size alone (e.g., patch size 16 at input resolution 576) preserves spatial locations while increasing segmentation feature resolution. As a result, although \method\ (medium) and \method-Seg (medium) both use 36 spatial locations, \method-Seg operates at lower input resolution, demonstrating that coupling segmentation feature resolution with input resolution shifts the Pareto-optimal operating point.

Further, we find that most Pareto-optimal \method \ models perform best with 2 windows, whereas LW-DETR achieves the best performance with 4 windows. We attribute this difference to how each architecture handles class tokens. LW-DETR’s CAEv2 backbone omits the class token, while \method's DINOv2 backbone relies on it as a key part of pre-training. To make windowed attention compatible with class tokens, we duplicate the class token for each window. During global attention, window-level class tokens attend to one another, while all other tokens attend to all class tokens. In practice, \method \ (nano), \method \ (small), and \method \ (medium) all use 2 windows, since duplicating class tokens for additional windows reduces runtime efficiency. As a result, unlike LW-DETR, \method \ does not benefit from scaling to 4 windows.


Lastly, we note that dataset characteristics influence optimal discovered architectures. We find that the optimal low latency models on RF100-VL tend to use fewer queries than the COCO models of equivalent latency. We attribute this to RF100-VL datasets having fewer objects per image than COCO.


\section{Visualizing Model Predictions}
\begin{figure}[h]
    \centering
    \includegraphics[width=0.9\linewidth]{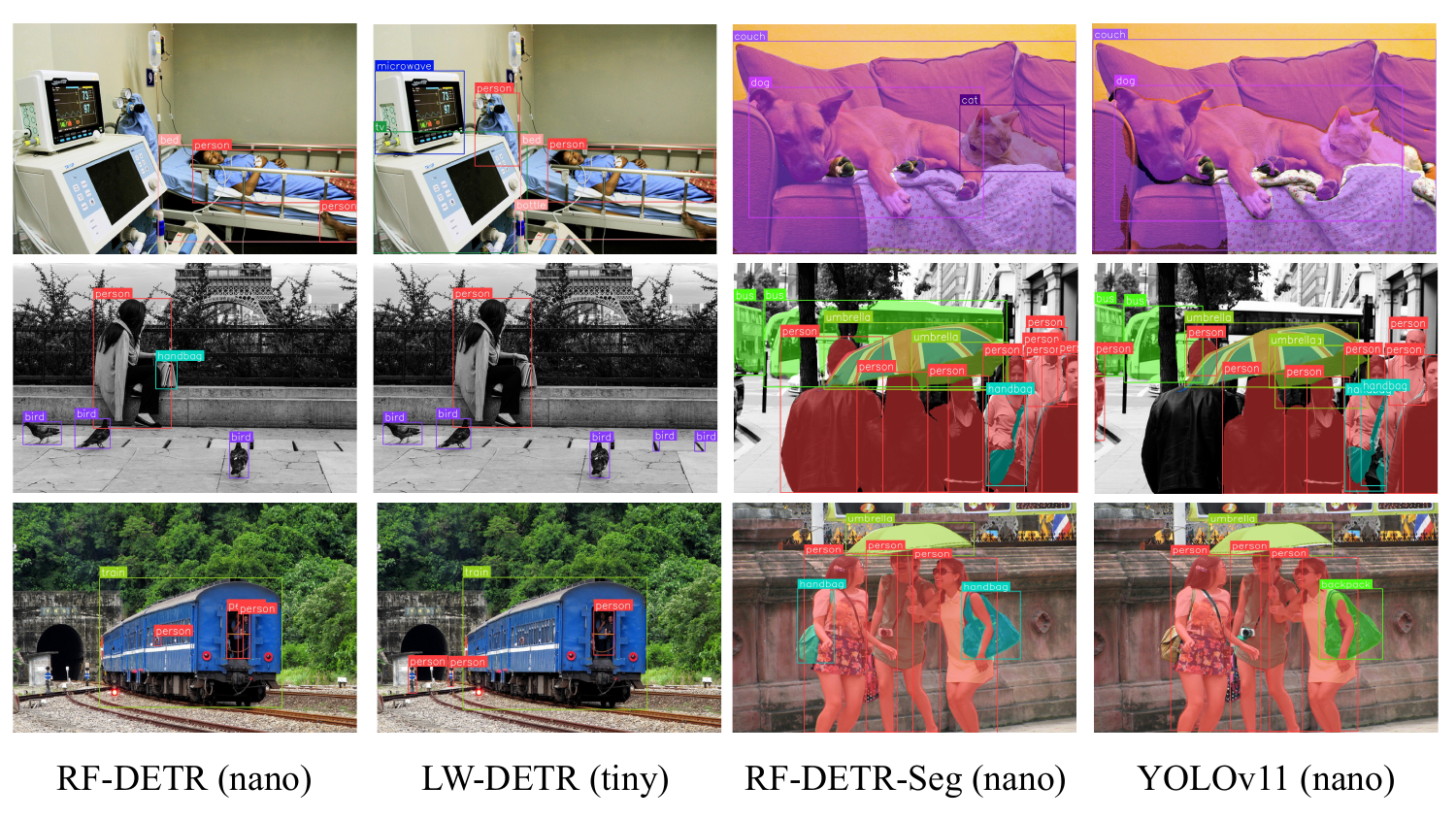}
    \caption{{\bf Visualizing Model Predictions}. On the left, we compare detections from \method \ (nano) and LW-DETR (tiny). On the right, we compare instance segmentation masks from \method-Seg (nano)  and YOLOv11 (nano)}
    \label{fig:visuals}
\end{figure}

We visualize model predictions from \method \ (nano) and relevant detection and segmentation baselines in Figure \ref{fig:visuals}. We find that \method \ (nano) predicts fewer false positives (e.g. mistaking {\tt sign post} for {\tt person}). Similarly, \method-Seg. (nano) predicts more precise object boundaries.

\end{document}